\documentclass[lettersize,journal]{IEEEtran}
\usepackage{amsmath,amsfonts}
\usepackage{algorithmic}
\usepackage{algorithm}
\usepackage{array}
\usepackage[caption=false,font=normalsize,labelfont=sf,textfont=sf]{subfig}
\usepackage{textcomp}
\usepackage{stfloats}
\usepackage{url}
\usepackage{verbatim}
\usepackage{graphicx}
\usepackage{cite}
\usepackage[x11names]{xcolor}
\usepackage[breaklinks=true,colorlinks,bookmarks=false,citecolor=SpringGreen4]{hyperref}
\usepackage{booktabs}       
\usepackage{multirow}
\usepackage{tablefootnote}
\usepackage{threeparttable}
\usepackage{bm}
\usepackage{color}
\usepackage{colortbl}
\usepackage{pifont}			
\usepackage{nicefrac}       
\usepackage{array}

\newcommand{\cmark}{\ding{51}}%
\newcommand{\xmark}{\ding{55}}%
\definecolor{Gray}{gray}{0.92}

\newcommand{\vs}{\textit{vs}.}
\newcommand{\etal}{\textit{et al}.}
\newcommand{\ie}{\textit{i}.\textit{e}.}
\newcommand{\eg}{\textit{e}.\textit{g}.}

\newtheorem{conclusion}{Conclusion}

\DeclareMathOperator{\Var}{\textit{Var}}
\DeclareMathOperator{\Cov}{\textit{Cov}}
\DeclareMathOperator{\Bias}{\textit{Bias}}

\def\etc{\emph{etc}}

\begin{document}

\title{Towards Better Accuracy-efficiency Trade-offs: Divide and Co-training}

\author{Shuai Zhao, Liguang Zhou, Wenxiao Wang, Deng Cai,~\IEEEmembership{Member,~IEEE}, \\
	Tin Lun LAM,~\IEEEmembership{Senior Member,~IEEE},
	and Yangsheng Xu,~\IEEEmembership{Fellow,~IEEE}
\thanks{
This work was supported by the Shenzhen Institute of Artificial Intelligence and  Robotics for Society under Grant AC01202101103.
This work was also supported in part by The National
Nature Science Foundation of China (Grant Nos: 62036009, 61936006).
(\emph{Corresponding author: Tin Lun Lam.})}
\thanks{Shuai Zhao, Liguang Zhou, Tin Lun LAM, and Yangsheng Xu
	are with the Shenzhen Institute of Artificial Intelligence and
	Robotics for Society (AIRS), The Chinese University of Hong Kong, Shenzhen,
	518129, Guangdong, China (e-mail: zhaoshuaimcc@gmail.com;
	liguangzhou@link.cuhk.edu.cn; tllam@cuhk.edu.cn; ysxu@cuhk.edu.cn).}
\thanks{Wenxiao Wang and Deng Cai are with the State Key Laboratory of CAD\&CG, Zhejiang
	University, Hangzhou 310027, China (email: wenxiaowang@zju.edu.cn; dengcai@zju.edu.cn).}
}

\maketitle

\begin{abstract}
The width of a neural network matters since increasing the width
will necessarily increase the model capacity.
However, the performance of a network does not improve linearly
with the width and soon gets saturated.
In this case, we argue that increasing the number of networks (ensemble)
can achieve better accuracy-efficiency trade-offs than purely increasing the width.
To prove it,
one large network is divided into several small ones
regarding its parameters and regularization components.
Each of these small networks has a fraction of the original one's parameters.
We then train these small networks together and make them see various 
views of the same data to increase their diversity.
During this co-training process,
networks can also learn from each other.
As a result, small networks can achieve better ensemble performance
than the large one with few or no extra parameters or FLOPs, \ie,
achieving better accuracy-efficiency trade-offs.
Small networks can also achieve faster inference speed
than the large one by concurrent running.
All of the above shows that the number of networks is a new dimension of model scaling.
We validate our argument with 8 different neural architectures on
common benchmarks through extensive experiments.
The code is available at \url{https://github.com/FreeformRobotics/Divide-and-Co-training}.
\end{abstract}


\section{Introduction} \label{sec_intro}
\begin{figure}[tbp]
	\centering
	\includegraphics[width=0.9\linewidth]{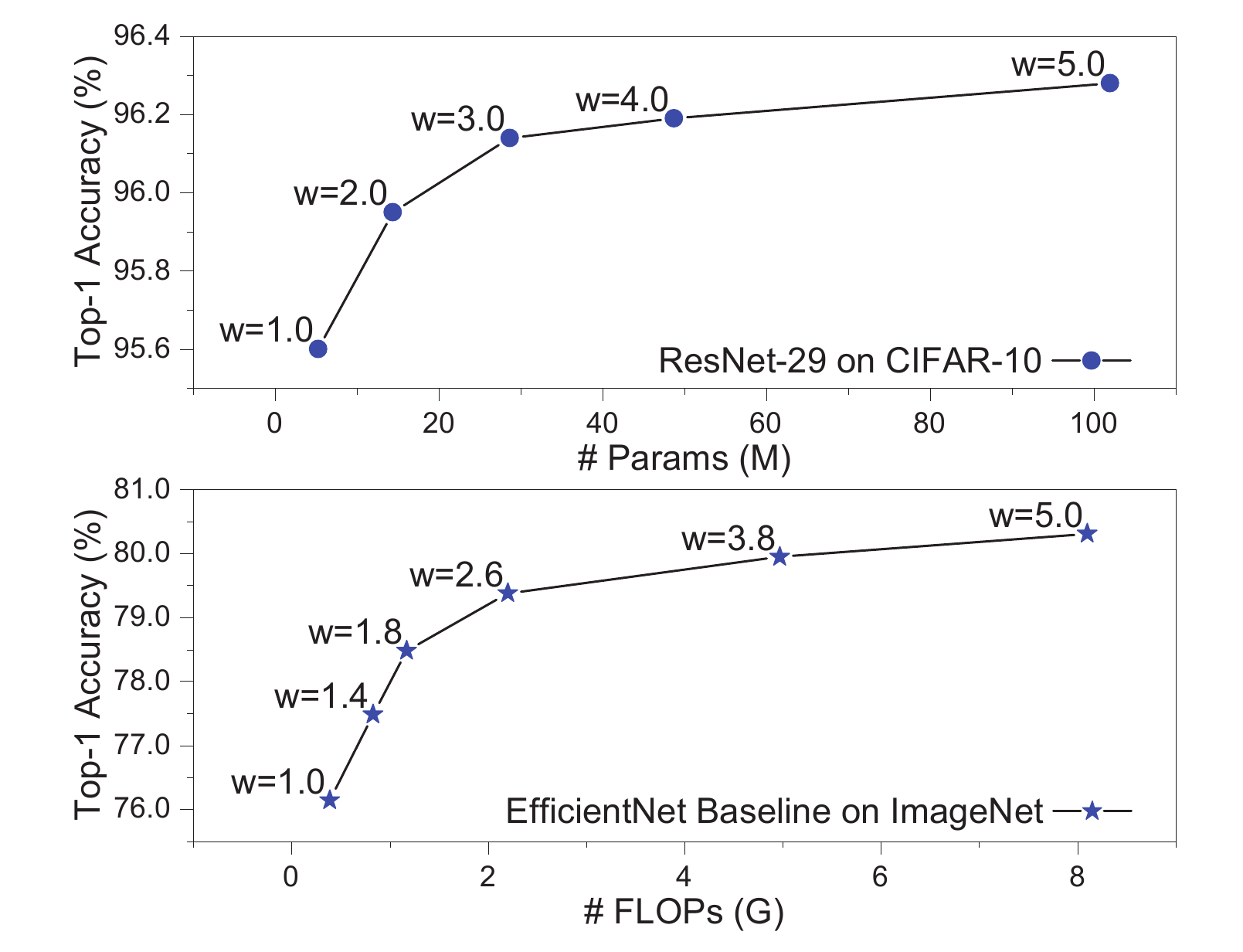}
	\caption{The width saturation of ResNeXt~\cite{2017_ResNeXt}
		and EfficientNet~\cite{2019_EfficientNet}.
		The gain does not match the expensive extra cost when the width ($w$) is large.
	}
	\label{fig_width_acc}
\end{figure}

\IEEEPARstart{I}{ncreasing} the width of neural networks to pursue better performance
is common sense in network engineering~\cite{2019_EfficientNet,2016_WRN, 2017_ResNeXt}.
However, the performance of networks does not improve linearly with the width.
As shown in Figure~\ref{fig_width_acc}, 
in the beginning, increasing width can gain promising improvement in accuracy;
at a later stage, the improvement becomes
slight and no longer matches the increasingly expensive cost.
For example, EfficientNet baseline ($w=5.0$, width factor) gains less than
$+0.5\%$ accuracy improvement compared to EfficientNet baseline ($w=3.8$) with nearly doubled 
FLOPs (floating-point operations).
We call this \emph{the width saturation of a network}. 
Increasing depth or resolution produces similar phenomena~\cite{2019_EfficientNet}.

Besides the width saturation,
we also observe that relatively small networks achieve close accuracies
to very wide networks, \ie, ResNet-29~($w=3.0$ \textit{v.s.} $w=5.0$)
and EfficientNet baseline ($w=2.6$ \textit{v.s.} $w=5.0$) in Figure~\ref{fig_width_acc}.
In this case, an interesting question arises,
can two small networks with a half width of a large one
achieve or even surpass the performance of the latter?
Firstly, ensemble is a practical technique and can improve
the generalization performance of individual neural networks
\cite{HuangLP0HW17, LiWD18, LiuY99, ROSEN96, abs-1904-05488}.    
Kondratyuk \etal~\cite{kondratyukWhenEnsemblingSmaller2020}
already demonstrate that the ensemble of several smaller networks is
more efficient than a large model in some cases.
Secondly, multiple networks can collaborate with their peers and
learning from each other during
training to achieve better individual or ensemble performance. 
This is verified by some deep mutual learning
\cite{2020_ICLR_MMT,yang2020mutualnet,2018_CVPR_DML}
and co-training~\cite{2018_QiaoSZWY18} works.

\begin{figure*}[tbp]
	\centering
	\small
	\includegraphics[width=0.84\linewidth]{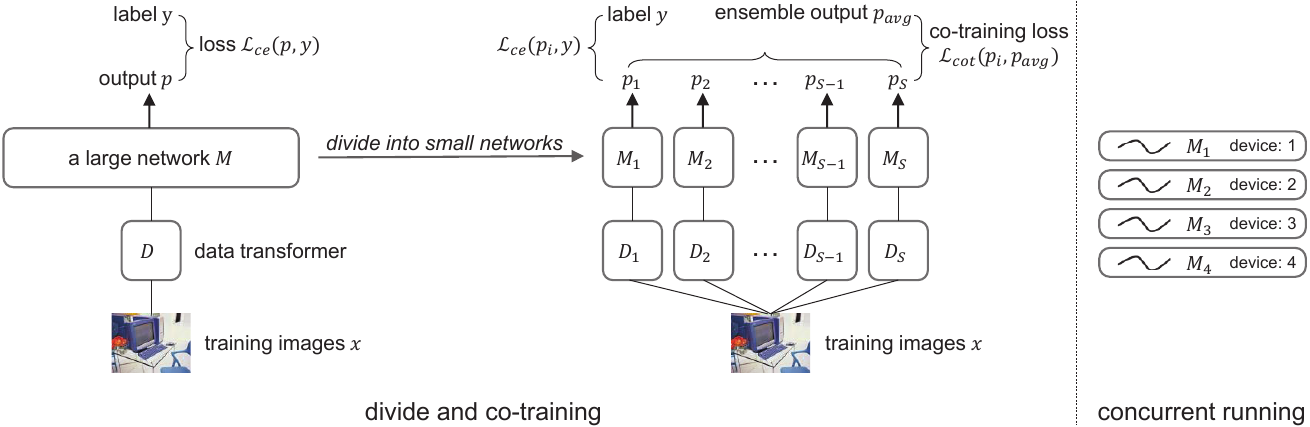} 
	\caption{Left: divide one large network into several small ones and co-train.
		Right: concurrent running of small networks on different devices during inference.
	}
	\label{fig_overall}
\end{figure*}

Based on the above analysis, we argue that increasing the number of networks (ensemble)
can achieve better accuracy-efficiency trade-offs than purely increasing the width.
A straightforward demonstration is given in this work: we divide one large network
into several pieces and show that these small networks can achieve
better ensemble performance than the large one with almost the same computation costs.

\begin{figure}[tbp]
	\centering
	\includegraphics[width=0.9\linewidth]{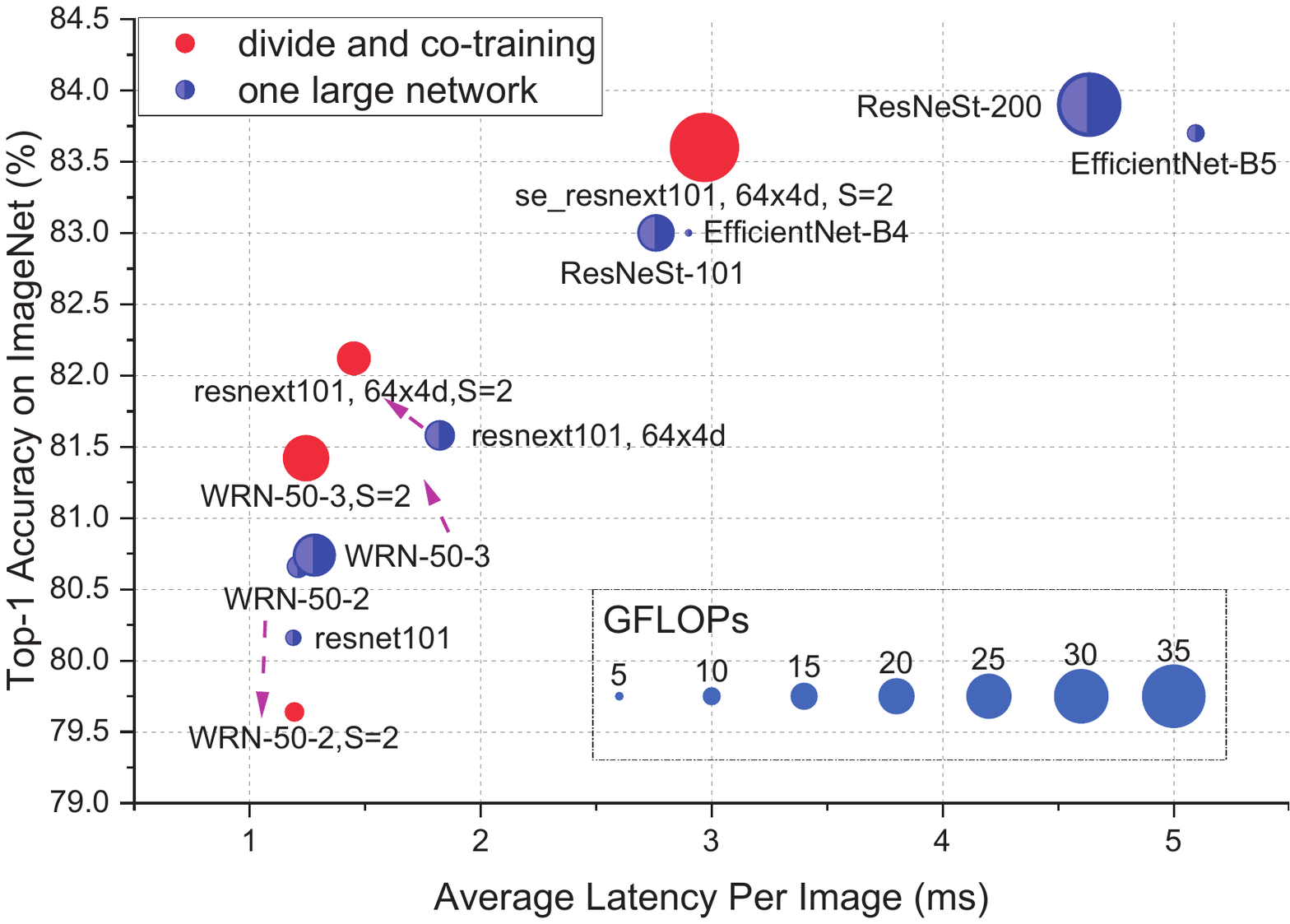}
	\caption{Networks and their inference latency.
		$S$ is the number of networks.
		Test on Tesla V100(s) with
		mixed precision~\cite{2018_AMP} and batch size 100. }
	\label{latency}
\end{figure}

The overall framework is shown in Figure~\ref{fig_overall}.
\emph{1) dividing:}
Given one large network, we first divide the network according
to its width, more precisely, the parameters or FLOPs of the network.
For instance, if we want to divide a network into two small networks, 
the number of the small one's parameters will become half of the original.
During this division process, the regularization components will
also be changed as the model capacity degrades.
Particularly, weight decay and drop layers will be divided
accordingly with the network width. 
\emph{2) co-training:}
After dividing,
small networks are trained with different views of the same
data~\cite{2018_QiaoSZWY18} to increase the diversity of networks
and achieve better ensemble performance~\cite{LiWD18}.
This is implemented by applying different data augmentation transformers in practice.
At the same time, small networks can also learn from
each other~\cite{2020_ICLR_MMT,2018_QiaoSZWY18,2018_CVPR_DML}
to further boost individual performance.
Thus we add Jensen-Shannon divergence among all predictions, \ie, 
co-training loss in Figure~\ref{fig_overall}.
In this way, one network can learn valuable knowledge about 
intrinsic object structure information from predicted posterior 
probability distributions of its peers~\cite{2015_HintonKD}.
\emph{3) concurrent running:}
Small networks can also achieve faster inference speed than the original one
through concurrent running on different devices when resources are sufficient.
Some different networks and their average latency~(inference speed) on
ImageNet~\cite{ImageNet} are shown in Figure~\ref{latency}.

We conduct extensive experiments with dividing and co-training
strategies for 8 different networks.
Generally, small networks after dividing accomplish better ensemble performance
than the original big network with similar parameters and FLOPs,
\ie, better accuracy-efficiency trade-offs are achieved.
We also reach competitive accuracy with state-of-the-art methods,
specifically, 98.31\% on CIFAR-10~\cite{cifar}, 89.46\% on CIFAR-100,
and 83.60\% on ImageNet~\cite{ImageNet}.
Furthermore, we validate our method for object detection
and demonstrate the generality of dividing and ensemble.
All evidence suggests that people should consider the number of networks
as a meaningful dimension of network engineering
besides commonly known width, depth, and resolution.

Our contributions are summarized as follows:
\begin{itemize}
	\item We post the claim that increasing the number of networks can achieve better
		accuracy-efficiency trade-offs than purely increasing the network width.
	\item We propose novel dividing strategies for 8 different networks regarding their
		parameters and regularizations to achieve better performance with similar costs. 
	\item Practical co-training techniques are developed to help small networks learn from their peers.
	\item We provide best practices and instructive conclusions for dividing and co-training
		via extensive experiments and thorough ablations on classification and object detection.
\end{itemize}

\section{Related works}
\noindent\paragraph*{Neural network architecture design}
Since the success of AlexNet~\cite{2012_conf/nips/KrizhevskySH12} on ImageNet~\cite{ImageNet},
deep learning methods dominate the field of computer vision, and
network engineering becomes a core topic.
Many excellent architectures emerged, \eg, NIN~\cite{lin2013network}, 
VGG~\cite{2014_VGGNet}, Inception~\cite{2015_SzegedyLJSRAEVR15}, 
ResNet~\cite{2016_ResNet}, Xception~\cite{2017_Xception}.
They explored different ways to design an effective and efficient model, \eg,
$1 \times 1$ convolution kernels,
stacked convolution layers with small kernels,
combination of different convolution and pooling operations,
residual connection,
depthwise separable convolution.
In recent years, neural architecture search (NAS) becomes more and more popular.
People hope to automatically learn or search for the best neural architectures
for certain tasks with machine learning methods.
We name a few here, 
reinforcement learning based NAS~\cite{2017_NAS},
progressive neural architecture search (PNASNet~\cite{2018_PNASNet}),
differentiable architecture search (DARTS~\cite{2019_ICLR_DARTS}), \etc.

\paragraph*{Implicit ensemble methods}
Ensemble methods use multiple learning algorithms to obtain better performance than
any of them.
Some layer splitting methods~\cite{2015_SzegedyLJSRAEVR15,2017_ResNeXt, zhang2020resnest}
adopt an implicitly "divide and ensemble" strategy, namely,
they divide a single layer in a model and then fuse their outputs to get better performance.
Dropout~\cite{journals/jmlr/SrivastavaHKSS14} can also be interpreted as
an implicit ensemble of multiple sub-networks within one full network.
Slimmable Network~\cite{iclr/YuYXYH19,2019_yuslimv2} derives several networks with different widths
from a full network and trains them in a parameter-sharing way to achieve adaptive
accuracy-efficiency trade-offs at runtime.
MutualNet~\cite{yang2020mutualnet} further trains these sub-networks mutually
to make the full network achieve better performance.
These methods get several dependent models after implicitly splitting,
while our methods obtain independent models
in terms of parameters after dividing. 
They are also compatible with our methods and can
be applied to any single model in our system.

\paragraph*{Collaborative learning}
Collaborative learning refers to a variety of educational approaches involving the
joint intellectual effort by students, or students and teachers together~\cite{smith1992collaborative}.
It was formally introduced in deep learning by~\cite{2018_NIPS_CDNN},
which was used to describe the simultaneous training of  
multiple classifier heads of a network.
Actually, according to its original definition, 
many works involving two or more models learning 
together can also be called collaborative learning, 
\eg, DML~\cite{2018_CVPR_DML}, co-training~\cite{1998_CO, 2018_QiaoSZWY18},
mutual mean-teaching (MMT)~\cite{2020_ICLR_MMT},  cooperative learning~\cite{2017_Coop},
knowledge distillation~\cite{2015_HintonKD}.
Their core idea is similar, \ie, 
enhancing the performance of one or all models by training them with 
some peers or teachers.
They inspire our co-training algorithm.

\section{Method}
\noindent
First of all, we recall the common framework of 
deep image classification.
Given a neural model $M$ and
$N$ training samples $\mathcal{X} = \{\bm{x}_n\}_{n=1}^{N}$ from $C$ classes,
the objective is cross entropy:
\begin{align}
\mathcal{L}_{ce}(p, y) = -\frac{1}{N} \sum_{n=1}^{N} \sum_{c=1}^{C} y_{n,c}\log(p_{n,c}), \label{softmax_ce}
\end{align}
where $y \in \{0,1\}$ is the ground truth label and
$p \in [0, 1]$ is the softmax normalized probability given by $M$.

\subsection{Division} \label{sec_divide}
\paragraph{Parameters}
In Figure~\ref{fig_overall},
we divide one large network $M$ 
into $S$ small networks $\{M_1, M_{2}, \ldots, M_{S} \}$.
The principle is keeping the metrics ---
the number of parameters or FLOPs roughly
unchanged before and after dividing.
$M$ is usually a stack of convolutional (conv.) layers.
Following the definition in PyTorch~\cite{NEURIPS2019_PyTorch},
its kernel size is $K \times K$, numbers of channels of input and output feature maps are
$C_{in}$ and $C_{out}$, and the number of groups is $d$, which means every
$\frac{C_{in}}{d}$ input channels are convolved with its own sets of filters,
of size $\frac{C_{out}}{d}$. In this case, the number of parameters and FLOPs are:
\begin{align}
\text{Params}:~
&K^2 \times \frac{C_{in}}{d} \times \frac{C_{out}}{d} \times d, \\
\text{FLOPs}:~
&(2 \times K^2 \times \frac{C_{in}}{d}  - 1 ) \times H \times W \times C_{out}, 
\label{flops}
\end{align} 
where $H \times W$ is the size of the output feature map and $-1$ occurs because 
the addition of $\frac{C_{in}}{d} \times K^2$ numbers only needs
$(\frac{C_{in}}{d} \times K^2 - 1)$
times operations. The bias is omitted for the sake of brevity.
For depthwise convolution~\cite{2018_mobilenetv2}, $d = C_{in}$.

Generally, $C_{out} = t_1 \times C_{in}$, where $t_1$ is a constant.
Therefore, if we want to divide a conv. layer by a factor $S$,
we just need to divide $C_{in}$ by $\sqrt{S}$:   
\begin{align}
\frac{K^2 \times C_{in} \times C_{out}}{S} \times \frac{1}{d} = 
K^2  \times t_1 \times (\frac{C_{in}}{ \sqrt{S} })^2 \times \frac{1}{d}.
\label{flops_divide}
\end{align} 
For instance, if we divide a bottleneck
$\Big[ \begin{smallmatrix}
1 \times 1, &64 \\
3 \times 3, &64 \\
1 \times 1, &256
\end{smallmatrix} \Big]$
in ResNet by 4, it becomes 4 small blocks
$\Big[ \begin{smallmatrix}
1 \times 1, &32 \\
3 \times 3, &32 \\
1 \times 1, &128
\end{smallmatrix} \Big]$.
Each small block has a quarter of the parameters or FLOPs
of the original block.

In practice, the numbers of output channels have the greatest common divisor (GCD).
The GCD of most ResNet
variants~\cite{2016_ResNet,2016_WRN,2017_ResNeXt,2018_CVPR_SENet} is the $C_{out}$ of
the first convolutional layer.
For other networks, like EfficientNet~\cite{2019_EfficientNet}, their GCD is a 
multiple of 8 or some other numbers.
In general, when we want to divide a network by $S$, we just need to find its GCD, 
and replace it with GCD$/\sqrt{S}$, then it is done.

For networks mainly consisted of group convolutions like ResNeXt~\cite{2017_ResNeXt},
we keep $\frac{C_{in}}{d}$ fixed, \ie, $C_{in} = t_2 \times d$,
where $t_2$ is a constant.
Namely, the number of channels per group is unchanged during dividing,
and the number of groups will be divided.
We substitute the $C_{in}$ in Eq.~\eqref{flops_divide} with the above
equation and get:
\begin{align}
K^2  \times t_1 \times t_2^2 \times \frac{d}{S}.
\end{align}
Then the division can be easily achieved by dividing $d$ by $S$.
This way is more concise than the square root operations.

For networks that have a constant global factor linearly related to the channel number, 
we simply divide the factor by $\sqrt{S}$,
\eg, the widen factor of WRN~\cite{2016_WRN},
the growth rate of DenseNet~\cite{2017_densenet},
and the additional rate of PyramidNet~\cite{2017_PyramidNet}.

\paragraph{Regularization}
After dividing, the model capacity degrades
and the regularization components in networks should change accordingly.
Under the assumption that the model capacity is linearly dependent with
the network width, we change the magnitude
of dropping regularization linearly.
Specifically, the dropping probabilities of dropout~\cite{journals/jmlr/SrivastavaHKSS14},
ShakeDrop~\cite{2019_shakedrop}, and stochastic depth~\cite{conf/eccv/HuangSLSW16}
are divided by $\sqrt{S}$ in our experiments.
As for weight decay~\cite{KroghH91_wd}, it is a little complex as
its intrinsic mechanism is still vague~\cite{2019_ICLR_ZhangWXG19,conf/nips/GolatkarAS19}.
We test some dividing manners and adopt two dividing strategies ---
no dividing and exponential dividing in this paper:
\begin{align}
wd^\star = wd \times \exp(\frac{1}{S} - 1.0), \label{wd_exp} 
\end{align}
where $wd$ is the original weight decay value and $wd^\star$ is the new value after dividing.
No dividing means the weight decay value keeps unchanged.
The above two dividing strategies are empirical criteria.
In practice, the best way now is trial and error.
Besides, we adopt exponential dividing rather than directly dividing the
$wd$ by $\sqrt{S}$ because the latter may lead to too small weight decay values
to maintain the generalization performance when $S$ is large.

The above dividing mechanism is usually not perfect, \ie,
the number of parameters of $M_i$ may not be exactly $1/S$ of $M$.
Firstly, $\sqrt{S}$ may not be an integer.
In this case, we round ${C_{in}}/{ \sqrt{S}}$ in Eq.~\eqref{flops_divide}
to a nearby even number.
Secondly, the division of the first layer and the
last fully-connected layer is not perfect because 
the input channel (color channels) and
output channel (number of object classes) are fixed.        

\paragraph{Concurrent running}
Small networks can also achieve faster inference
speed than the large network by concurrent running on different
devices as shown in Figure~\ref{fig_overall}. 
Typically, a device is an NVIDIA GPU. 
Theoretically, if one GPU has enough processing units, \eg,
streaming multiprocessor,
small networks can also run concurrently within one GPU
with multiple CUDA Streams~\cite{cuda_toolkit}.
However, we find this is impractical on a Tesla V100 GPU in experiments.
One small network is already able to occupy most of the computational resources,
and different networks can only run in sequence.
Therefore, we only discuss the multiple devices fashion.


\subsection{Co-training}
The generalization error of neural network ensemble~(NNE)
can be interpreted as
\emph{Bias-Variance-Covariance Trade-off}~\cite{LiuY99, ROSEN96}.
Let $f$ be the function model learned, $g(\bm{x})$ be the target function,
and $f_{en}(\bm{x}; \mathcal{X})=\frac{1}{S}\sum_{i=1}^{S}f_i(\bm{x}; \mathcal{X})$.
Then the expected mean-squared ensemble error is:
\begin{small}
	\begin{align}
	&\mathbb{E}_{\mathcal{X}}[\big(f_{en}(\bm{x}; \mathcal{X}) - g(\bm{x})\big)^2] = 
	\big(\mathbb{E}_{\mathcal{X}}[f_{en}(\bm{x}) - g(\bm{x})]\big)^2  \nonumber \\
	&+ \mathbb{E}_{\mathcal{X}}\big[\frac{1}{S^2}\sum\nolimits_{i=1}^{S}\big(f_i(\bm{x}; \mathcal{X}) - \mathbb{E}_{\mathcal{X}}[f_i(\bm{x}; \mathcal{X})]\big)^2 \big] \nonumber \\
	&+ \mathbb{E}_{\mathcal{X}}\big[\frac{1}{S^2}\sum\nolimits_{i=1}^{S}\sum\nolimits_{j\ne i}^{S}
	\big(f_i(\bm{x}; \mathcal{X}) - \mathbb{E}_{\mathcal{X}}[f_i(\bm{x}; \mathcal{X})]\big) 
	\nonumber \\
	&\times \big(f_j(\bm{x}; \mathcal{X}) - \mathbb{E}_{\mathcal{X}}[f_j(\bm{x}; \mathcal{X})]\big)\big], ~\label{var-cov-bias}
	\end{align} 
\end{small}where the first term is the square bias~($\Bias^2$) of the combined system,
the second and third terms are the variance~($\Var$) and covariance~($\Cov$)
of the outputs of individual networks.
A clean form is
$\mathbb{E}[ \frac{1}{S}\Var + (1 - \frac{1}{S})\Cov + \Bias^2]$.
Data noise is omitted here.
The detailed proof is given by Ueda \etal~\cite{UedaN96}.

\paragraph{Different initialization and data views}
Increasing the diversity of networks without
increasing $\Var$ or $\Bias$ can decrease
the correlation ($\Cov$) between networks.
To this end, small networks are
initialized with different weights~\cite{LiWD18}.
Then, when feeding the training data, we apply different data transformers $D_i$
on the same data for different networks
as shown in Figure~\ref{fig_overall}.   
In this way, $\{M_1, M_2, \ldots, M_{S} \}$ are trained on different views
$\{D_1(\bm{x}), D_{2}(\bm{x}),\ldots, D_{S}(\bm{x})\}$ of $\bm{x}$.
In practice, different data views are generated by the randomness
of data augmentation. Besides the commonly used random resize,
random crop, and random flip,
we introduce random erasing~\cite{zhong2020random} and
AutoAugment~\cite{2019_autoaugment} policies.
AutoAugment has 14 image transformation operations, 
\eg, shear, translate, rotate, and auto contrast.
It searches tens of different policies which
are consisted of two operations and randomly
chooses one policy during the data augmentation process.
By applying these random data augmentations,
we can guarantee that $D_i(\bm{x})$ produces
different views of $\bm{x}$ across multiple runs.     
%
\paragraph{Co-training loss}
Knowledge distillation and
DML~\cite{2018_CVPR_DML} show
that one network can boost its performance by learning
from a teacher or cohorts.
Namely, a deep ensemble system can reduce its $\Bias$ in this way.
Besides, following the co-training assumption~\cite{1998_CO},
small networks are expected to have consistent predictions on  
$\bm{x}$ although they see different views of $\bm{x}$.
From the perspective of Eq.~\eqref{var-cov-bias},
this can reduce the variance of networks and avoid poor performance
caused by overly decorrelated networks~\cite{ROSEN96}.
Therefore, we adopt Jensen-Shannon~(JS) divergence among predicted probabilities
as the co-training objective~\cite{2018_QiaoSZWY18} :
\begin{align}
\mathcal{L}_{cot}(p_1, p_{2}, \ldots, p_{S}) = 
H(\frac{1}{S}\sum_{i=1}^{S} p_i) - \frac{1}{S} \sum_{i=1}^{S} H(p_i),
\label{cot_loss}
\end{align}
where $p_i$ is the estimated probability of $M_i$,
and $H(p) = \mathbb{E} [-\log (p)]$ is the Shannon entropy of the 
distribution of $p$. Through this co-training manner,
one network can learn valuable information from its peers, which
defines a rich similarity structure over objects.
For example, a model classifies an object as \textit{Chihuahua}
may also give high confidence about \textit{Japanese spaniel} since
they are both dogs~\cite{2015_HintonKD}.
DML uses the Kullback-Leibler~(KL)
divergence between predictions of every two networks.
Their purpose is similar to ours.

The overall objective function is
a combination of the
classification losses of small networks and the co-training loss:
\begin{align}
\mathcal{L}_{all} = \sum_{i=1}^{S} \mathcal{L}_{ce}(p_i, y) + 
\lambda_{cot} \mathcal{L}_{cot}(p_1, p_{2}, \ldots, p_{S}),
\label{overall_loss}
\end{align}
where $\lambda_{cot} = 0.5$ is a weight factor of $\mathcal{L}_{cot}(\cdot)$
and it is chosen by cross-validation.
At the early stage of training, the outputs
of networks are full of randomness, 
so we adopt a warm-up scheme for
$\lambda_{cot}$~\cite{2017_conf/iclr/LaineA17,2018_QiaoSZWY18}.
Specifically, we use a linear scaling up strategy when
the current training epoch is less than a certain number ---
40 and 60 for CIFAR and ImageNet, respectively.    
$\lambda_{cot} = 0.5$ is also an equilibrium point between
learning diverse networks and producing consistent predictions.
During inference, we average the outputs before
softmax layers as the final ensemble output.

\section{Experiment}

\subsection{Experimental setup} \label{sec_exp_setup}

\noindent\textbf{Datasets~} 
We adopt CIFAR-10, CIFAR-100~\cite{cifar}, and
ImageNet 2012~\cite{ImageNet} datasets.
CIFAR-10 and CIFAR-100 contain
50K training and 10K test RGB images of size 32$\times$32,
labeled with 10 and 100 classes, respectively.
ImageNet 2012 contains 1.28 million training images
and 50K validation images from 1000 classes.
For CIFAR and ImageNet, crop size is 32 and 224,
batch size is 128 and 256, 
respectively.

\noindent\textbf{Learning rate and training epochs~}
We apply warm up and
cosine learning rate policy~\cite{2017_PriyaGoyal,2018_bags_of_tricks}.
If the initial learning rate is $lr$ and current epoch is $epoch$,
for the first $\textit{slow\_epoch}$ steps, the learning rate is 
$lr \times \frac{\textit{epoch}}{\textit{slow\_epoch}}$;
for the rest epochs, the learning rate is
$0.5 \times lr \times(1 + \cos ( \pi \times 
\frac{\textit{epoch} - \textit{slow\_epochs}}{\textit{max\_epoch} - \textit{slow\_epoch}} ) )$.
Generally, $lr$ is 0.1; $\{\textit{max\_epoch}, \textit{slow\_epoch} \}$ is
\{300, 20\} and \{120, 5\} for CIFAR and ImageNet, respectively.
Models before and after dividing are trained for the same epochs.


\begin{table}[b]
	\centering
	\caption{
		Influence of various experiment settings on CIFAR-100. step-lr means step learning rate policy~\cite{2016_ResNet}.
		When the erasing probability $p_e=1.0$, random erasing acts like cutout~\cite{2017_cutout}.
	}
	\label{tab1_various_settings}
	\resizebox{\linewidth}{!}{%
		\begin{tabular}{l|ccccccc}
			\toprule
			\rowcolor{Gray}
			Method
			& step-lr  				& cos-lr		
			& rand. erasing 		& mixup 		
			& AutoAug.
			& Top-1 err. (\%) 		\\
			\midrule
			\multirow{6}{*}{\shortstack{ResNet-110~\cite{2016_ResNetPreAct} \\
					original: 26.88\%}}
			
			& \cmark				& ~	
			& ~							& ~					
			& ~								
			& 24.71 $\pm$ 0.22	\\
			
			~	
			& ~   						& \cmark		
			& ~ 						& ~ 				
			& ~ 									
			& 24.15 $\pm$ 0.07				\\
			
			~
			& ~  						& \cmark 				
			& \cmark, $p_e=1.0$		& 			
			& ~									
			& 23.43 $\pm$ 0.01				\\
			
			~
			& ~  						& \cmark  				
			& \cmark, $p_e=0.5$		& 			
			& ~								
			& 23.11 $\pm$ 0.29 				\\
			
			~
			& ~  						& \cmark  				
			& \cmark, $p_e=0.5$		& \cmark, $\lambda=0.2$ 	 		
			& ~									
			& 21.22 $\pm$ 0.28				\\
			
			~
			& ~  						& \cmark 				
			& \cmark, $p_e=0.5$		& \cmark, $\lambda=0.2$ 	 		
			& \cmark							
			& \textbf{19.19} $\pm$ 0.23				\\
			\bottomrule	
	\end{tabular}}
\end{table}
\noindent\textbf{Data augmentation~}
Random crop and resize, random left-right flipping, 
AutoAugment~\cite{2019_autoaugment}, random erasing~\cite{zhong2020random},
and mixup~\cite{2018_mixup} are used during training.
Label smoothing~\cite{2016_Rethinking} is only applied when 
training on ImageNet.

\noindent\textbf{Weight decay~}
Generally, weight decay is \textit{1e-4}.
For \{EfficientNet-B3, ResNeXt-29, WRN-28-10, WRN-40-10\} on
CIFAR datasets, it is \textit{5e-4}.
Bias and parameters of batch normalization~\cite{2015_BN} are left undecayed.

Besides, we use kaiming weight initialization~\cite{2015_DelvingRec}.
The optimizer is nesterov~\cite{nesterov1983method} accelerated SGD with a momentum 0.9.
ResNet variants adopt the modifications introduced in~\cite{2018_bags_of_tricks},
\ie, replacing the first $7 \times 7$ conv. with
three consecutive $3 \times 3$ conv. and 
put an $2 \times 2$ average pooling layer before $1\times 1$ conv. 
when there is a need to downsample.
The code is implemented in PyTorch~\cite{NEURIPS2019_PyTorch}.
Influence of some settings is shown in Table~\ref{tab1_various_settings}.

\begin{table*}[!t]
	\centering
	\caption{
		Results of Top-1 error (\%) on CIFAR-100.
		The last three rows are trained for 1800 epochs.
		$S$ is the number of small networks after dividing. 
		Train from scratch, no extra data.
		Weight decay value keeps unchanged except \{WRN-40-10, 1800 epochs\},
		which applies Eq.~\eqref{wd_exp}.
		The maximal reduction of error of different networks is shown in {\color{blue} blue}.
	}
	\label{tab2_cifar100_results}
	\resizebox{0.94\linewidth}{!}{%
		\begin{threeparttable}
			\begin{tabular}{l|c||ccc||ccc||cccccc}
				\toprule
				
				\rowcolor{Gray}
				& original
				& \multicolumn{3}{c||}{re-implementation}	
				& \multicolumn{3}{c||}{(\# Networks) $\bm{S=2}$} 
				& \multicolumn{3}{c}{(\# Networks) $\bm{S=4}$}	\\
				
				\rowcolor{Gray}
				\multirow{-2}{*}{\centering Method}
				&  error
				&  error
				&  MParams
				&  GFLOPs		
				&  error
				&  MParams
				&  GFLOPs					
				&  error
				&  MParams
				&  GFLOPs			\\
				\midrule		
				
				ResNet-110~\cite{2016_ResNetPreAct}
				& 26.88
				& 18.96
				& 1.17
				& 0.17		
				& \textbf{18.32}{\color{blue}$_{(0.64~\downarrow)}$}
				& 1.33
				& 0.20
				& 19.56$_{(0.63~\uparrow)}$
				& 1.21
				& 0.18			\\				
				
				ResNet-164~\cite{2016_ResNetPreAct}
				& 24.33
				& 18.38
				& 1.73
				& 0.25		
				& \textbf{17.12}{\color{blue}$_{(1.26~\downarrow)}$}
				& 1.96
				& 0.29
				& \textbf{18.05}$_{(0.33~\downarrow)}$
				& 1.78
				& 0.26			\\
				
				SE-ResNet-110~\cite{2018_CVPR_SENet}
				& 23.85
				& 17.91
				& 1.69
				& 0.17
				& \textbf{17.37}{\color{blue}$_{(0.54~\downarrow)}$}
				& 1.89
				& 0.20
				& 18.33$_{(0.42~\uparrow)}$
				& 1.70
				& 0.18			\\			
				
				SE-ResNet-164~\cite{2018_CVPR_SENet}
				& 21.31
				& 17.37
				& 2.51
				& 0.26	
				& \textbf{16.31}{\color{blue}$_{(1.06~\downarrow)}$}
				& 2.81
				& 0.29
				& \textbf{17.21}$_{(0.16~\downarrow)}$
				& 2.53
				& 0.27		\\		
				
				EfficientNet-B0~\cite{2019_EfficientNet}$^\dagger$
				& -
				& 18.50
				& 4.13
				& 0.23		
				& \textbf{18.20}$_{(0.30~\downarrow)}$
				& 4.28
				& 0.24
				& \textbf{17.85}{\color{blue}$_{(0.65~\downarrow)}$}
				& 4.52
				& 0.30			\\
				
				EfficientNet-B3~\cite{2019_EfficientNet}$^\dagger$
				& -
				& 18.10
				& 10.9
				& 0.60		
				& \textbf{17.00}$_{(1.10~\downarrow)}$
				& 11.1
				& 0.60
				& \textbf{16.62}{\color{blue}$_{(1.48~\downarrow)}$}
				& 11.7
				& 0.65			\\

				WRN-16-8~\cite{2016_WRN}
				& 20.43
				& 18.69
				& 11.0
				& 1.55		
				& \textbf{17.37}$_{(1.32~\downarrow)}$
				& 12.4
				& 1.75
				& \textbf{17.07}{\color{blue}$_{(1.62~\downarrow)}$}
				& 11.1
				& 1.58			\\
				
				ResNeXt-29, 8$\times$64d~\cite{2017_ResNeXt}
				& 17.77
				& 16.43
				& 34.5
				& 5.41
				& \textbf{14.99}$_{(1.44~\downarrow)}$
				& 35.4
				& 5.50
				& \textbf{14.88}{\color{blue}$_{(1.55~\downarrow)}$}
				& 36.9
				& 5.67
				\\				
				
				WRN-28-10~\cite{2016_WRN}
				& 19.25
				& 15.50
				& 36.5
				& 5.25		
				& \textbf{14.48}$_{(1.02~\downarrow)}$
				& 35.8
				& 5.16
				& \textbf{14.26}{\color{blue}$_{(1.24~\downarrow)}$}
				& 36.7
				& 5.28			\\				
				
				WRN-40-10~\cite{2016_WRN}
				& 18.30
				& 15.56
				& 55.9
				& 8.08		
				& \textbf{14.28}$_{(1.28~\downarrow)}$
				& 54.8
				& 7.94
				& \textbf{13.96}{\color{blue}$_{(1.60~\downarrow)}$}
				& 56.0
				& 8.12			\\
				\midrule
				
				DenseNet-BC-190~\cite{2017_densenet}
				& 17.18
				& 14.10
				& 25.8
				& 9.39		
				& \textbf{12.64}$_{(1.46~\downarrow)}$
				& 25.5
				& 9.24
				& \textbf{12.56}{\color{blue}$_{(1.54~\downarrow)}$}$^\sharp$
				& 26.3
				& 9.48			\\
				
				PyramidNet-272~\cite{2017_PyramidNet}+ShakeDrop~\cite{2019_shakedrop}
				& 14.96
				& 11.02
				& 26.8
				& 4.55		
				& \textbf{10.75}$_{(0.27~\downarrow)}$
				& 28.9
				& 5.24
				& \textbf{10.54}{\color{blue}$_{(0.48~\downarrow)}$}$^\sharp$
				& 32.8
				& 6.33			\\
				
				WRN-40-10~\cite{2016_WRN}
				& 18.30
				& 16.02
				& 55.9
				& 8.08		
				& \textbf{14.09}$_{(1.93~\downarrow)}$
				& 54.8
				& 7.94
				& \textbf{13.10}{\color{blue}$_{(2.92~\downarrow)}$}$^\sharp$
				& 56.0
				& 8.12			\\
				\bottomrule	
			\end{tabular} 
			{\footnotesize \begin{tablenotes}
					\item[$\dagger$] When training on CIFAR-100, the stride of EfficientNet at 
					the stage 4\&7 is set to be 1. Original EfficientNet on CIFAR-100 is pre-trained while we
					train it from scratch.
					\item[$\sharp$] These results rank \#2, \#6, and \#7 on public leaderboard
					\href{https://paperswithcode.com/sota/image-classification-on-cifar-100}{paperswithcode.com/sota/image-classification-on-cifar-100}
					(without extra training data)
					at the time of submission.
			\end{tablenotes}}
		\end{threeparttable}
	}
\end{table*}

\subsection{Results on CIFAR and ImageNet dataset}
\subsubsection{CIFAR-100} \label{sec_cifar_100}
In Table~\ref{tab2_cifar100_results},
dividing and co-training achieve consistent improvements
with similar or even fewer parameters or FLOPs.
Additional cost occurs because the division of a
network is not perfect, as mentioned in Sec.~\ref{sec_divide}.
Three conclusions can be drawn from these results 7 networks.
\begin{conclusion} \label{conclu_number}
\textit{Increasing the number, width, and depth of networks together
is more efficient and effective than purely increasing the width or depth.}
\end{conclusion}

For all networks in Table~\ref{tab2_cifar100_results},
dividing and co-training gain promising improvement.
We also notice,
ResNet-110~($S=2$) $>$ ResNet-164,
SE-ResNet-110~($S=2$) $>$ SE-ResNet-164,
EfficientNet-B0~($S=4$) $>$ EfficientNet-B3,
and WRN-28-10~($S=4$) $>$ WRN-40-10
with fewer parameters,
where $>$ means the former has better performance.
By contrast, the latter is deeper or wider.
Besides, with wider or deeper networks,
dividing and co-training can gain more improvement,
\eg, ResNet-110~(+0.64) \textit{vs.} ResNet-164~(+\textbf{1.26}), 
SE-ResNet-110~(+0.54) \textit{vs.} SE-ResNet-164~(+\textbf{1.06}),
WRN-28-10~(+1.24) \textit{vs.} WRN-40-10~(+\textbf{1.60}), and
EfficientNet-B0~(+0.65) \textit{vs.} EfficientNet-B3~(+\textbf{1.48}).
All of the above demonstrates the superiority of increasing the number of networks.
It is also clear that increasing the number, width, and depth of networks together
is a better choice than purely scaling up one single dimension during model engineering.

\begin{conclusion} \label{conclu_perform}
\textit{The ensemble performance is closely related to individual performance.}
\end{conclusion}

\begin{figure}[!t]
	\centering
	\includegraphics[width=0.9\linewidth]{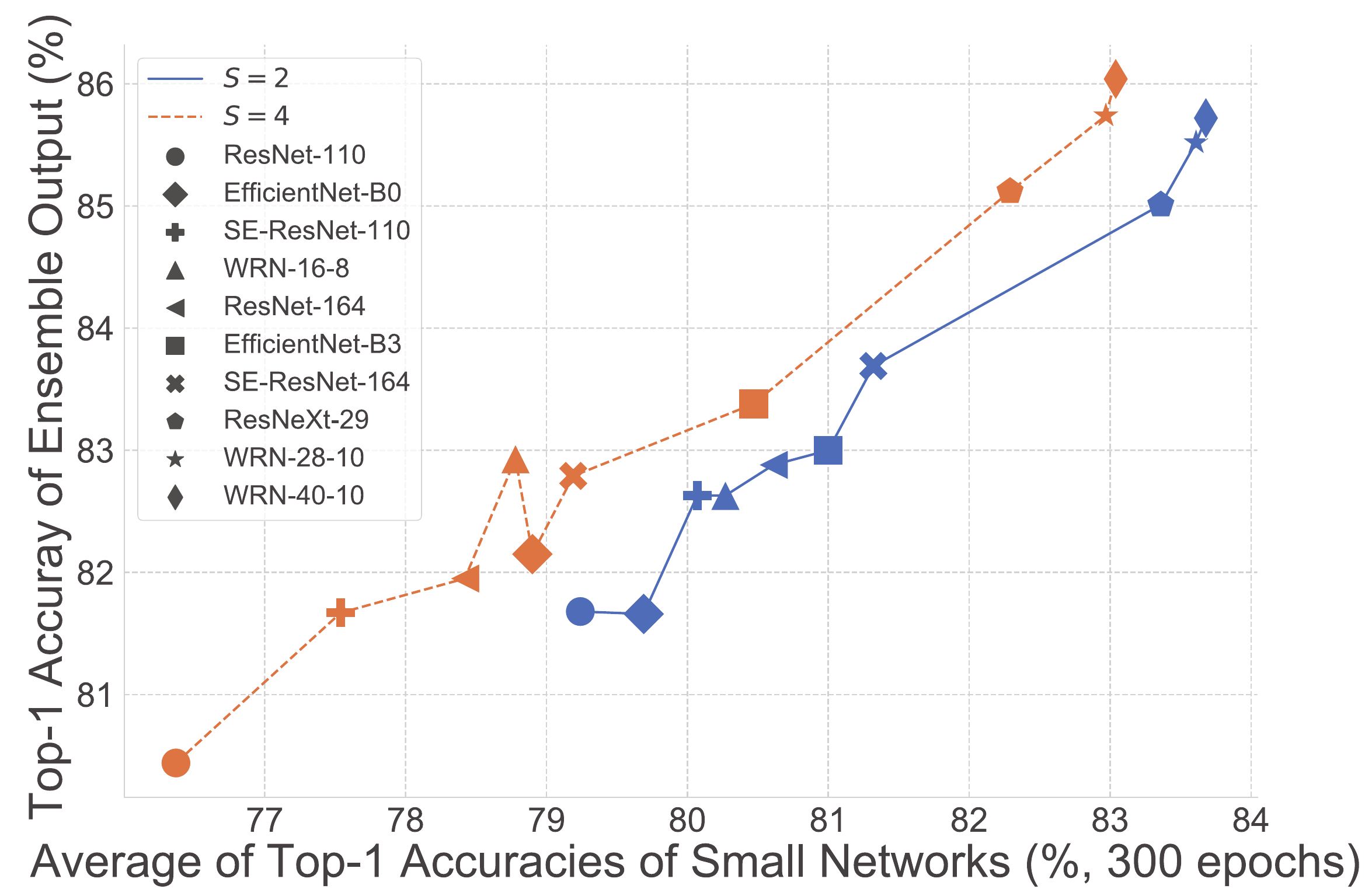}
	\caption{The relationship between the  numerical average of accuracies
		and ensemble accuracy of small networks on CIFAR-100.}
	\label{average_acc}
\end{figure}

The relationship between the average accuracy and
ensemble accuracy is shown in Figure~\ref{average_acc}.
When calculating the average accuracy,
we separately calculate the accuracies of small networks and average them.
From the big picture, the average accuracy is positively correlated with
the ensemble accuracy.
The higher the average accuracy, the better the ensemble performance.
This coincides with the Eq.~\eqref{var-cov-bias} because the stronger the
individual networks the smaller the $\Bias$.

At the same time, we note that there is a big gap between 
the average accuracy and ensemble accuracy.
The ensemble accuracy is higher than the average accuracy by
a large margin.
This gap may be filled by two factors according to Eq.~\eqref{var-cov-bias}:
the decayed variance of individual networks and learned diverse networks
during the co-training process.

\begin{conclusion}  \label{conclu_necessary}
	\textit{A necessary width/depth of networks matters.}
\end{conclusion}

In Table~\ref{tab2_cifar100_results},
ResNet-110~($S=4$) and SE-ResNet-110~($S=4$) get a drop in performance,
0.63\%$\downarrow$ and 0.42\%$\downarrow$, respectively.
In Figure~\ref{average_acc}, these two networks obtain the first two lowest
average accuracies.
If we take one step further to look at the architecture of these small networks,
we will find they have input channels 
$[8, 16, 32]$ at the first layer of their three blocks.
These networks are too thin compared to the original one with  
$[16, 32, 64]$ channels.
In this case, the magnitude of the decayed variance of individual networks
is smaller than the magnitude of gained accuracy~(decayed bias)
from increasing the channels~(width) of a single network.
Consequently, individual networks cannot gain satisfying accuracy
and lead to inferior ensemble performance as conclusion~\ref{conclu_perform} said.
This demonstrates that a necessary width is essential.
The conclusion also applies to PyramidNet-272,
which gets a relatively slight improvement with
dividing and co-training as its base channel is small, \ie, 16.
Increasing the width of networks is still effective when the width is very small.

\begin{table}[!t]
	\centering
	\caption{
		Results of Top-1 error (\%) on CIFAR-10.
		$S$ is the number of small networks after dividing.
		Divide weight decay as Eq.~\eqref{wd_exp}.
		No extra data and train from scratch.
		Methods compared here are AutoAugment~\cite{2019_autoaugment},
		RandAugment~\cite{cubuk2019randaugment},
		Fixup-init~\cite{2019_Fixup},
		Cutout~\cite{2017_cutout},
		Mixup~\cite{2018_mixup},
		ShakeDrop~\cite{2019_shakedrop},
		and
		Fast AutoAugment~\cite{2019_fastAA}.
	}
	\label{tab3_cifar10_results}
	\resizebox{0.96\linewidth}{!}{%
		\begin{threeparttable}
			\begin{tabular}{lcccc}
				\toprule
				\rowcolor{Gray}
				Method
				& \multicolumn{2}{c}{error}
				& {MParams}	
				& {GFLOPs}			\\

				\midrule
				
				ResNet-164~\cite{2016_ResNetPreAct}
				& \multicolumn{2}{c}{5.46}
				& 1.73	
				& 0.25			\\	
				
				SE-ResNet-164~\cite{2018_CVPR_SENet}
				& \multicolumn{2}{c}{4.39}
				& 2.51	
				& 0.26			\\
				
				WRN-28-10~\cite{2016_WRN}
				& \multicolumn{2}{c}{4.00}
				& 36.5	
				& 5.25			\\	
				
				ResNeXt-29, 8$\times$64d~\cite{2017_ResNeXt}
				& \multicolumn{2}{c}{3.65}
				& 34.5	
				& 5.41			\\
				
				WRN-28-10~\cite{2016_WRN, 2019_autoaugment}
				& \multicolumn{2}{c}{2.68}
				& 36.5	
				& 5.25			\\
				
				WRN-28-10~\cite{2016_WRN,cubuk2019randaugment}
				& \multicolumn{2}{c}{2.70}
				& 36.5	
				& 5.25			\\						
				
				WRN-40-10~\cite{2016_WRN, 2019_Fixup, 2018_mixup, 2017_cutout}
				& \multicolumn{2}{c}{2.30}
				& 55.9
				& 8.08			\\				
				
				Shake-Shake 26 2$\times$96d~\cite{2017_shakeshake,2019_autoaugment}
				& \multicolumn{2}{c}{2.00}
				& 26.2	
				& 3.78			\\
				
				Shake-Shake 26 2$\times$96d~\cite{2017_shakeshake, cubuk2019randaugment}
				& \multicolumn{2}{c}{2.00}
				& 26.2	
				& 3.78			\\				
				
				PyramidNet-272~\cite{2017_PyramidNet,2019_shakedrop,2019_fastAA}
				& \multicolumn{2}{c}{1.70}
				& 26.2	
				& 4.55			\\
				\midrule
				\rowcolor{Gray}
				
				~
				& \multicolumn{2}{c}{epochs}
				&
				& \\

				\rowcolor{Gray}
				\multirow{-2}{*}{\centering re-implementation}
				& 300
				& 1800	
				& \multirow{-2}{*}{\centering MParams}
				& \multirow{-2}{*}{\centering GFLOPs}			\\
				\midrule
				
				SE-ResNet-164
				& 2.98
				& 2.19
				& 2.49					
				& 0.26			\\
				
				ResNeXt-29, 8$\times$64d
				& 2.69
				& 2.23
				& 34.4	
				& 5.40			\\

				WRN-28-10
				& 2.28
				& 2.41
				& 36.5	
				& 5.25			\\
				
				WRN-40-10
				& 2.33
				& 2.19
				& 55.8	
				& 8.08			\\
				
				Shake-Shake 26 2$\times$96d
				& 
				& 2.00
				& 26.2	
				& 3.78			\\
				
				
				\midrule
				
				SE-ResNet-164,~$S=2$
				& \textbf{2.84}
				& 2.20
				& 2.81					
				& 0.29			\\
				
				ResNeXt-29, 8$\times$64d,~$S=2$ 
				& \textbf{2.12}
				& \textbf{1.94}
				& 35.1	
				& 5.49			\\	
				
				WRN-28-10,~$S=2$ 
				& \textbf{2.06}
				& \textbf{1.81}
				& 35.8	
				& 5.16			\\	
				
				WRN-28-10,~$S=4$ 
				& \textbf{2.01}
				& ~\textbf{1.68}$^\sharp$
				& 36.5	
				& 5.28			\\
				
				
				WRN-40-10,~$S=4$
				& \textbf{2.01}
				& ~\textbf{1.62}$^\sharp$
				& 55.9	
				& 8.12			\\
				
				Shake-Shake 26 2$\times$96d,~$S=2$
				& 
				& \textbf{1.75}
				& 23.3	
				& 3.38			\\	
				
				Shake-Shake 26 2$\times$96d,~$S=4$
				& 
				& ~\textbf{1.69}$^\sharp$
				& 26.3	
				& 3.82			\\
				\bottomrule	
			\end{tabular} 
			{\footnotesize \begin{tablenotes}
					\item[$\sharp$] These results rank \#6, \#7, and \#8 on public leaderboard
					 \href{https://paperswithcode.com/sota/image-classification-on-cifar-10}{paperswithcode.com/sota/image-classification-on-cifar-10} (without extra training data)
					at the time of submission.
			\end{tablenotes}}
		\end{threeparttable}
	}
\end{table}

In Table~\ref{tab2_cifar100_results}, 
ResNet-164~($S=4$) and SE-ResNet-164~($S=4$) still get improvements, although
their performance is not as good as applying ($S=2$).  
In Figure~\ref{average_acc}, small networks of ResNet-164 and 
SE-ResNet-164 with ($S=4$) also get better performance than
ResNet-110 and SE-ResNet-110 with ($S=4$), respectively.
This reveals that a necessary depth can help guarantee the
model capacity and achieve better ensemble performance.
In a word, after dividing, 
the small networks should maintain a necessary width or depth to guarantee their model
capacity and achieve satisfying ensemble performance.
This is consistent with some previous works.
Kondratyuk \etal~\cite{kondratyukWhenEnsemblingSmaller2020}
show that the ensemble of some small models with limited depths and widths cannot
achieve promising ensemble performance compared to some large models.

From conclusion 1 -- 3, 
we can learn some best practices about effective model scaling.
When the width/depth of networks is small,
increasing width/depth can still get substantial improvement.
However, when width or depth becomes large and 
increasing them yields little gain~(Figure~\ref{fig_width_acc}),
it is more effective to increase the number of networks.
This comes to our core conclusion --- \emph{Increasing the
number, width, and depth of networks together
is more efficient and effective than purely scaling up one dimension
of them.}

\subsubsection{CIFAR-10}
Results on the CIFAR-10 are shown in Table~\ref{tab3_cifar10_results}.
Although the Top-1 accuracy on CIFAR-10 is already very high,
dividing and co-training can also achieve significant improvement.
It is worth noting \{WRN-28-10, epoch 1800\} gets worse performance
than \{WRN-28-10, epoch 300\}, namely,
WRN-28-10 may overfit on CIFAR-10 when trained for more epochs.
In contrast, dividing and co-training can help WRN-28-10 get consistent improvement 
with increasing training epochs.
This is because we can learn diverse networks from the data.
In this way, even if one model overfits,
the ensemble of different models can also make a comprehensive and correct prediction.
From the perspective of Eq.~\eqref{var-cov-bias}, diverse networks reduce the $\Var$
or achieve a negative $\Cov$.
The conclusion also applies to
\{WRN-40-10, 300 epochs\} and \{WRN-40-10, 1800 epochs\} on CIFAR-100.
This shows ensembles of neural networks are not only more accurate but also
more robust than individual networks. 

\begin{table*}[htbp]
	\centering
	\small
	\caption{
		Single crop results on ImageNet 2012 validation set.
		No extra data and train from scratch.
		$S$ is the number of small networks after dividing.
		Acc. of $M_i$ is the accuracy of small networks.
		Only WRN applies Eq.~\eqref{wd_exp}; others keep weight decay unchanged.
	}
	\label{tab_imagenet_results}
	\resizebox{0.84\linewidth}{!}{%
		\begin{threeparttable}
			\begin{tabular}{lccccc}
				\toprule
				\rowcolor{Gray}
				Method
				& Top-1~/~Top-5 Acc.
				& MParams~/~GFLOPs 		
				& Crop~/~Batch~/~Epochs
				& Acc. of $M_i$\\
				\midrule
				
				WRN-50-2~\cite{2016_WRN}
				& 78.10~~~/~~~93.97
				& 68.9~~~/~~~12.8			
				& 224~~/~~256~~/~120 
				& \multirow{10}{*}{---}	\\
				
				ResNeXt-101,~64$\times$4d~\cite{2017_ResNeXt}
				& 79.60~~~/~~~94.70
				& 83.6~~~/~~~16.9			
				& 224~~/~~256~~/~120 \\
				
				ResNet-200 + AutoAugment~\cite{2019_autoaugment}
				& 80.00~~~/~~~95.00
				& 64.8~~~/~~~16.4				
				& 224~~/~4096~/~270 \\
				
				
				SENet-154~\cite{2018_CVPR_SENet}
				& 82.72~~~/~~~96.21
				& 115.0~~/~~~42.3
				& 320~~/~1024~/~100  \\	
				
				SENet-154 + MultiGrain~\cite{2019arXivMultiGrain}
				& 83.10~~~/~~~96.50
				& 115.0~~/~~~83.1
				& 450~~/~~512~~/~120 \\			
				
				PNASNet-5~$(N=4, F=216)$~\cite{2018_PNASNet}$^\dagger$
				& 82.90~~~/~~~96.20
				& 86.1~~~/~~~25.0
				& 331~~/~1600~/~312 \\	
				
				AmoebaNet-C~$(N=6, F=228)$~\cite{2019_amobanet}$^\dagger$
				& 83.10~~~/~~~96.30
				& 155.3~~/~~~41.1
				& 331~~/~3200~/~100 \\
				
				ResNeSt-200~\cite{zhang2020resnest}
				& ~\textbf{83.88}~~~/~~~~~---~~~
				& 70.4~~~/~~~35.6
				& 320~~/~2048~/~270 \\
				
				EfficientNet-B6~\cite{2019_EfficientNet}
				& \textbf{84.20}~~~/~~~\textbf{96.80}
				& 43.0~~~/~~~19.0
				& 528~~/~4096~/~350 \\
				
				EfficientNet-B7~\cite{2019_EfficientNet}
				& \textbf{84.40}~~~/~~~\textbf{97.10}
				& 66.7~~~/~~~37.0
				& 600~~/~4096~/~350 \\			
				
				\midrule		
				
				WRN-50-2~(re\_impl.)
				& 80.66~~~/~~~95.16
				& 68.9~~~/~~~12.8
				& 224~~/~~256~~/~120 
				& \multirow{4}{*}{---}\\
				
				WRN-50-3~(re\_impl.)
				& 80.74~~~/~~~95.40
				& 135.0~~/~~~23.8			
				& 224~~/~~256~~/~120  \\
				
				ResNeXt-101,~64$\times$4d~(re\_impl.)
				& 81.57~~~/~~~95.73
				& 83.6~~~/~~~16.9				
				& 224~~/~~256~~/~120 \\
				
				EfficientNet-B7~(re\_impl.)$^\sharp$
				& 81.83~~~/~~~95.78
				& 66.7~~~/~~~10.6				
				& 320~~/~~256~~/~120 
				\\			
				\midrule	
				
				WRN-50-2,~$S=2$
				& 79.64~~~/~~~94.82
				& 51.4~~~/~~~10.9	
				& 224~~/~~256~~/~120 
				& 78.68, 78.66\\
				
				WRN-50-3,~$S=2$
				& \textbf{81.42}~~~/~~~\textbf{95.62}
				& 138.0~~/~~~25.6				
				& 224~~/~~256~~/~120 
				& 80.32, 80.24 \\
				
				ResNeXt-101,~64$\times$4d,~$S=2$
				& \textbf{82.13}~~~/~~~\textbf{95.98}
				& 88.6~~~/~~~18.8				
				& 224~~/~~256~~/~120 
				& 81.09, 81.02 \\
				
				EfficientNet-B7,~$S=2$
				& \textbf{82.74}~~~/~~~\textbf{96.30}
				& 68.2~~~/~~~10.5	
				& 320~~/~~256~~/~120 
				& 81.39, 81.83 \\

				EfficientNet-B7,~$S=4$
				& \textbf{82.75}~~~/~~~\textbf{96.22}
				& 70.9~~~/~~~12.6	
				& 320~~/~~256~~/~120 
				& 80.49, 80.55, 80.56, 80.48 \\

				
				SE-ResNeXt-101,~64$\times$4d,~$S=2$
				& \textbf{83.60}~~~/~~~\textbf{96.69}
				& 98.0~~~/~~~38.2				
				& 320~~/~~128~~/~350
				& 82.43, 82.46 \\

				\bottomrule	
			\end{tabular} 
			\begin{tablenotes}
				{\footnotesize 
					\item[$\dagger$] PNASNet and AmoebaNet use 100 P100 workers. On each worker, batch size is 16 or 32.
				}
			\end{tablenotes}
		\end{threeparttable}
	}
\end{table*}

\subsubsection{ImageNet}
Results on ImageNet are shown in Table~\ref{tab_imagenet_results}.
All experiments on ImageNet are conducted with mixed precision~\cite{2018_AMP}.
WRN-50-2 and WRN-50-3 are 2$\times$ and 3$\times$ wide as ResNet-50, respectively.
The results on ImageNet validate our argument again --- 
Increasing the number, width, and depth of networks together
is more efficient and effective than purely scaling one dimension of width, depth, and resolution.
Specifically,
EfficientNet-B7~(84.4\%, 66M, crop 600, wider, deeper)
\vs
EfficientNet-B6~(84.2\%, 43M, crop 528), 
WRN-50-3~(80.74\%, 135.0M)
\vs
WRN-50-2~(80.66\%, 68.9M),
the former
only produces 0.2\%$\uparrow$
and 0.08\%$\uparrow$ gain of accuracy, respectively.
However, the cost is nearly doubled.
This shows that increasing width or depth 
can only yield little gain when the model is already very large,
while it brings out an expensive extra cost. 
In contrast, increasing the number of networks rewards
with much more tangible improvement.

\subsection{Discussion}
\paragraph{influence of dividing regularization terms}
The influence of dividing the regularization components
(weight decay and dropping layers) is shown in Table~\ref{tab_divwd}.
The results partially support our assumption: small models generally
need less regularization.
There are also some counterexamples: dividing the weight decay of
WRN-16-8 does not work. Possibly \textit{1e-4} is a small number for
WRN-16-8 on CIFAR-100, and it should not be further divided.
It is worth noting that \textit{5e-4} and \textit{1e-4} are already
appropriate weight decay values selected by cross-validation for WRN-28-10 and WRN-16-8,
respectively.

\begin{table}[tbp]
	\centering
	\caption{
		Influence of dividing manners of weight decay and dropping layers.
		$p_{shake}$ is the initial dropping probability of ShakeDrop.
	}
	\label{tab_divwd}
	\resizebox{0.96\linewidth}{!}{%
		\begin{tabular}{llcc}
			\toprule
			\rowcolor{Gray}
			~
			& ~ 
			& \multicolumn{2}{c}{Top-1 err.~(\%)}	\\	
			
			\rowcolor{Gray}
			\multirow{-2}{*}{Method} 
			& \multirow{-2}{*}{\textit{wd}}
			& CIFAR-10	
			& CIFAR-100
			\\
			
			\midrule
			
			WRN-28-10
			& \textit{5e-4}
			& 2.28
			& 15.50
			\\
			
			WRN-28-10,~$S=2$
			& \textit{5e-4}
			& 2.15
			& 14.48	
			\\	
			
			WRN-28-10,~$S=2$
			& \textit{2.5e-4}
			& 2.15
			& 14.43	
			\\
			
			WRN-28-10,~$S=2$
			& Eq.~\eqref{wd_exp}
			& \textbf{2.06}
			& \textbf{14.16}  
			\\	
			
			\midrule
			WRN-28-10,~$S=4$
			& \textit{5e-4}
			& 2.36
			& 14.26	
			\\	
			
			WRN-28-10,~$S=4$
			& \textit{1.25e-4}
			& \textbf{2.00}
			& 14.79
			\\
			
			WRN-28-10,~$S=4$
			& Eq.~\eqref{wd_exp}
			& 2.01
			& \textbf{14.04} 
			\\
			\midrule
			WRN-16-8
			& \textit{1e-4}
			& \multirow{4}{*}{---}
			& 18.69 \\
			
			WRN-16-8,~$S=2$	
			& \textit{1e-4} 
			& ~
			& \textbf{17.37} \\
			
			WRN-16-8,~$S=2$
			& \textit{0.5e-4}
			& ~
			& 18.11		\\
			
			WRN-16-8,~$S=2$
			& Eq.~\eqref{wd_exp}
			& ~	
			& 17.77		\\
			
			\midrule
			\midrule
			\rowcolor{Gray}
			& ~   
			& \multicolumn{2}{c}{Top-1 err.~(\%)} \\			
			
			\rowcolor{Gray}
			\multirow{-2}{*}{Method}
			& \multirow{-2}{*}{$p_{shake}$ }
			& CIFAR-10	
			& CIFAR-100
			\\
			
			\midrule
			
			PyramidNet-272~+~ShakeDrop
			& 0.5
			& \multirow{3}{*}{---} 
			& 11.02
			\\
			
			PyramidNet-272~+~ShakeDrop,~$S=2$
			& 0.5
			& 
			& 11.15	
			\\	
			
			PyramidNet-272~+~ShakeDrop,~$S=2$
			& $0.5/\sqrt{2}$
			& 
			& \textbf{10.75}	
			\\			
			\bottomrule	
	\end{tabular} }
\end{table}

\paragraph{different ensemble methods}
Besides the averaging ensemble manner,
we also test max ensemble,
\ie, use the most confident prediction of small networks as the final prediction,
and the geometric mean of model predictions~\cite{kondratyukWhenEnsemblingSmaller2020}:
\begin{align}
p = \big(\prod_i^S p_i\big)^{\frac{1}{S}}. \label{eq_geo_mean}
\end{align}
Results are shown in Table~\ref{tab_ensenmble manners}.
Simple averaging is the most effective way among the test methods in most cases.

\begin{table}[tbp]
	\centering
	\caption{
		Influence of different ensemble manners.
		Softmax~(\cmark) means we ensemble the results after softmax operation.
		Generally, we do ensemble operations without softmax, \ie, softmax~(\xmark)
	}
	\label{tab_ensenmble manners}
	\resizebox{0.96\linewidth}{!}{%
		\begin{tabular}{lcccc}
			\toprule
			\rowcolor{Gray}
			~
			& ~
			& ~  
			& \multicolumn{2}{c}{Top-1 err.~(\%)}			
			\\	
			\rowcolor{Gray}
			\multirow{-2}{*}{Method} 
			& \multirow{-2}{*}{\textit{softmax}}
			& \multirow{-2}{*}{\textit{ensemble}} 
			& CIFAR-10	
			& CIFAR-100		
			\\
			
			\midrule
			
			\multirow{4}{*}{{WRN-28-10,~$S=4$}}
			& \cmark
			& max
			& 1.85
			& 15.04
			\\
			
			~
			& \cmark
			& avg.
			& 1.77	
			& 14.45
			
			\\	
			
			~
			& \xmark
			& max
			& 1.90			
			& 15.27
			\\
			
			~
			& \xmark
			& avg.
			& \textbf{1.68} 
			& \textbf{14.26}
			\\	
			
			~
			& \cmark
			& geo. mean
			& \textbf{1.68} 	
			& \textbf{14.26} \\
			
			\midrule
			\multirow{5}{*}{{ResNeXt-29, 8$\times$64d,~$S=2$}}
			& \cmark
			& max 		
			& 1.94
			& 15.58
			\\
			~
			& \cmark
			& avg.
			& \textbf{1.93}
			& 15.08
			\\
			
			~
			& \xmark
			& max
			& 2.01
			& 15.78
			\\
			~
			& \xmark
			& avg.	
			& 1.94
			& \textbf{14.99}	\\
			~
			& \cmark
			& geo. mean
			& 1.94	
			& 15.00 \\
			\midrule
			\midrule
			\rowcolor{Gray}
			& ~
			& ~
			& \multicolumn{2}{c}{ImageNet Acc.~(\%)}	
			\\
			\rowcolor{Gray}
			\multirow{-2}{*}{Method} 
			& \multirow{-2}{*}{\textit{softmax}}
			& \multirow{-2}{*}{\textit{ensemble}} 
			& Top-1	
			& Top-5
			\\
			\midrule
			%
			%
			\multirow{5}{*}{{ResNeXt-101,~64$\times$4d,~$S=2$}}	
			& \cmark
			& max 		
			& 81.92
			& 95.82
			\\
			~
			& \cmark
			& avg.
			& 82.03
			& 95.91
			\\
			
			~
			& \xmark
			& max
			& 81.78
			& 95.80
			\\
			
			~
			& \xmark
			& avg.	
			& \textbf{82.13}
			& \textbf{95.98}	
			\\	
			
			~
			& \cmark
			& geo. mean
			& 82.12	
			& 95.93 \\
			\bottomrule	
	\end{tabular} }
\end{table}

\paragraph{influences of different co-training components}
The influences of dividing and ensemble,
different data views, and various values of 
weight factor $\lambda_{cot}$ of co-training loss
in Eq.~\eqref{overall_loss} are shown in Table~\ref{tab4_abla_cifar100}.
The contribution of dividing and ensemble is the most significant,
\ie, 1.10\%$\uparrow$ at best.
Using different data transformers (0.30\%$\uparrow$)
and co-training loss (0.41\%$\uparrow$) can also help the model improve
performance.
Besides, the effect of co-training is more significant when there are more networks,
\ie, 0.18\%$\uparrow$ ($S=2$) \textit{vs}. 0.41\%$\uparrow$ ($S=4$).
Considering the baseline is very strong (see Table~\ref{tab1_various_settings}),
their improvement is also significant.

We do not divide the large network into 8, 16, or more small networks.
Firstly, the large the $S$, the thinner the small networks. 
As conclusion \ref{conclu_necessary} in Sec. CIFAR-100~\ref{sec_cifar_100} said,
after dividing,
the small networks may be too thin to guarantee a necessary model capacity
and satisfying ensemble performance \---\ unless the original network is very large.
Secondly, as mentioned in Sec. Division~\ref{sec_divide}, small networks run in sequence during training,
and training of 8 or 16 small networks may cost too much time.
The implementation of an asynchronous training system needs further hard work.
Theoretically, one can abandon the co-training part to achieve an easy one
with some sacrifice of performance.

\begin{table}[tbp]
	\centering
	\caption{
		Influence of co-training components on CIFAR-100.
	}
	\label{tab4_abla_cifar100}
	\resizebox{0.96\linewidth}{!}{%
		\begin{tabular}{l|ccc|ccc}
			\toprule
			\rowcolor{Gray}	
			& \multicolumn{3}{c}{(\# Networks) $\bm{S=2}$}
			& \multicolumn{3}{c}{(\# Networks) $\bm{S=4}$}		\\
			
			\rowcolor{Gray}
			\multirow{-2}{*}{\centering Method}
			& diff. views 
			& $\lambda_{cot}$			
			& Top-1 err. (\%)		
			& diff. views
			& $\lambda_{cot}$		
			& Top-1 err. (\%)	 \\
			\midrule		
			
			\multirow{5}{*}{\shortstack{WRN-16-8 \\ 
					original: 20.43\% \\
					re-impl.: 18.69\%}}	
			& \xmark
			& ~
			& 17.85
			& \xmark
			& ~
			& 17.59			\\				
			
			~	
			& \cmark
			& ~
			& 17.55
			& \cmark
			& ~
			& 17.48			\\	
			
			~	
			& \cmark
			& 0.1
			& 17.50
			& \cmark
			& 0.1
			& 17.33			\\	
			
			~	
			& \cmark
			& 0.5
			& \textbf{17.37}
			& \cmark
			& 0.5
			& \textbf{17.07}	 \\	
			
			~	
			& \cmark
			& 1.0
			& 17.48
			& \cmark
			& 1.0
			& 17.53		\\			
			\bottomrule	
	\end{tabular} }
\end{table}

In this work, we just use a simple co-training method
and make no further exploration in this direction.
There do exist some other more complex co-training or mutual learning methods.
For example, MutualNet~\cite{yang2020mutualnet} derives several networks with various widths
from a single full network, feeds them images at different resolutions,
and trains them together in a weight-sharing way to boost the performance of the full network.
We mainly focus on validating the effectiveness and efficiency of
increasing the number of networks,
more complex ensemble and co-training methods are left as future topics.

\begin{table}[tbp]
	\centering
	\caption{
		Average inference runtime (ms) per sample
		and average cosine similarity of outputs of small networks on CIFAR-100.
		Test on RTX 2080Ti.
		seq. and para. means that several networks run in sequence and parallel,
		respectively. 
		Smaller time is in {\color{blue} blue}.
	}
	\label{runtime}
	\resizebox{\linewidth}{!}{
		\begin{tabular}{l|c|ccc|ccc}
			\toprule
			\rowcolor{Gray}
			& original~($\bm{S=1}$)
			& \multicolumn{3}{c}{(\# Networks) $\bm{S=2}$}
			& \multicolumn{3}{c}{(\# Networks) $\bm{S=4}$}		\\
			
			\rowcolor{Gray}
			& ~
			& \multicolumn{2}{c}{runtime}
			& ~
			& \multicolumn{2}{c}{runtime} 
			& ~
			\\
			
			\rowcolor{Gray}
			\multirow{-3}{*}{\centering Method}
			& \multirow{-2}{*}{\centering runtime}
			& seq.
			& para.
			& \multirow{-2}{*}{\centering cos. sim.}
			& seq. 
			& para. 
			& \multirow{-2}{*}{\centering cos. sim.}\\
			
			\midrule		
			%
			
			ResNet-110
			& {\color{blue} 0.24}
			& 0.41
			& 0.52	
			& 0.84 
			& 0.83
			& 1.08
			& 0.85			\\
			
			ResNet-164
			& {\color{blue} 0.32}
			& 0.66
			& 0.71	
			& 0.83 
			& 1.25
			& 1.27
			& 0.83			\\
			EfficientNet-B0
			& {\color{blue} 0.27}
			& 0.37
			& 0.39	
			& 0.79
			& 0.68
			& 0.78
			& 0.79			\\
			EfficientNet-B3
			& {\color{blue} 0.51}
			& 0.68
			& 0.63	
			& 0.85
			& 1.00
			& 1.15
			& 0.85			\\
			
			ResNeXt-29, 8$\times$64d
			& 1.13
			& 1.20
			& {\color{blue} 0.72}	
			& 0.89
			& 1.49
			& {\color{blue} 0.74}
			& 0.89			\\
			
			WRN-16-8
			& {\color{blue} 0.24}
			& 0.29
			& {\color{blue} 0.24}	
			& 0.80
			& 0.38
			& 0.38
			& 0.78			\\
			
			WRN-28-10
			& 0.60
			& 0.67
			& {\color{blue} 0.45}	
			& 0.87
			& 0.91
			& {\color{blue} 0.55}
			& 0.87			\\				
			
			WRN-40-10
			& 0.87
			& 0.98
			& {\color{blue} 0.61}	
			& 0.88
			& 1.31
			& {\color{blue} 0.70}
			& 0.80			\\	
			\bottomrule	
	\end{tabular}}
\end{table}

\begin{figure}[t]
	\centering
	\includegraphics[width=0.9\linewidth]{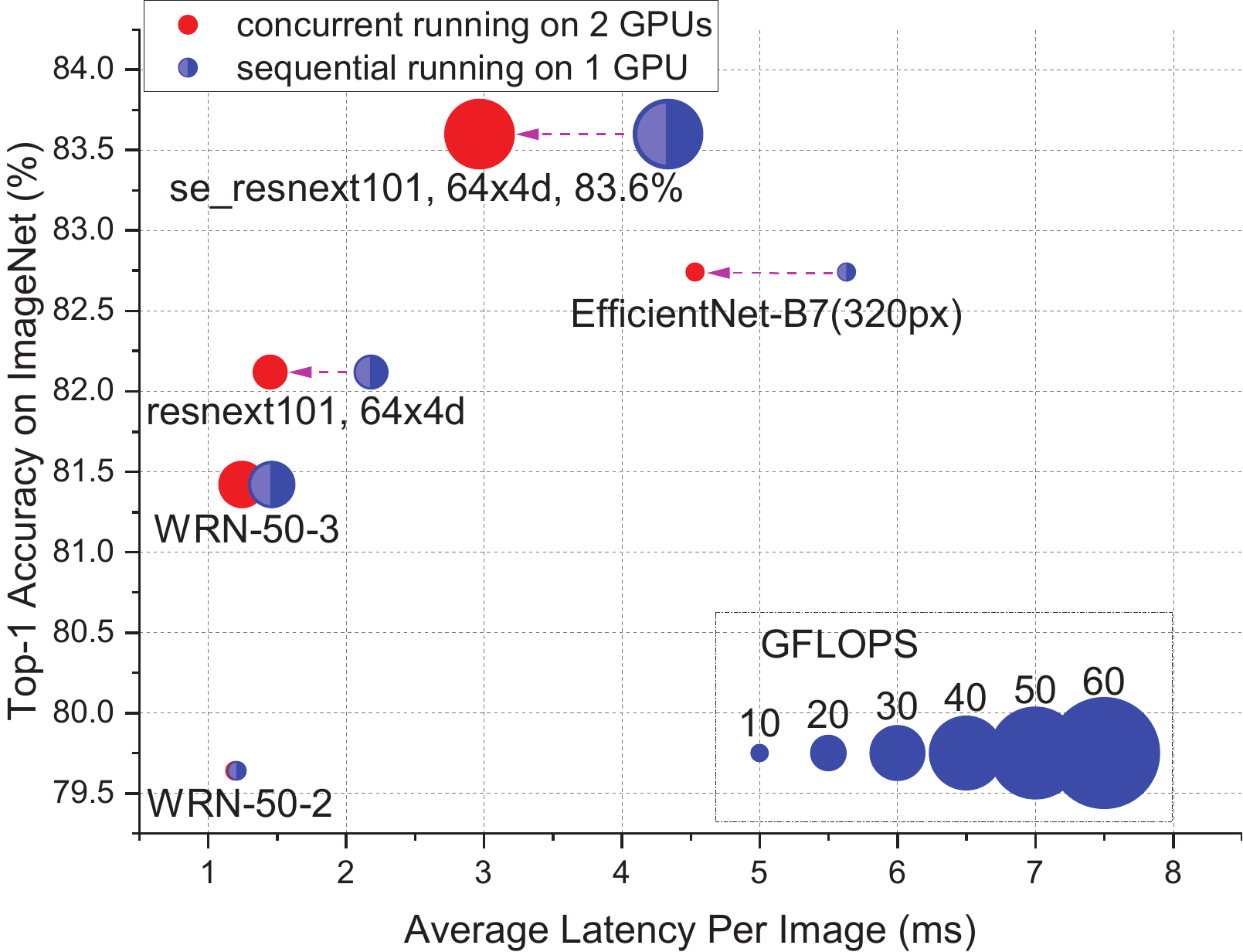}
	\caption{The difference of inference latency between 
		sequential and concurrent running.
		Test on Tesla V100(s) with
		mixed precision and batch size 100. }
	\label{latency_comp}
\end{figure}

\paragraph{runtime and correlation of small networks after dividing}
Runtime and cosine similarity of outputs of different networks
are shown in Table~\ref{runtime}.
Corresponding accuracy and FLOPs of these networks can be found in
Table~\ref{tab2_cifar100_results}.
For big networks, concurrent running (small networks run in parallel)
can get obvious speedup at runtime compared to sequential running.
For small networks, additional data loading, data pre-processing (still runs in sequence) and
data transferring (CPU$\Leftrightarrow$GPU, GPU$\Leftrightarrow$GPU) will occupy
most of the time. 
The runtime is also related to the intrinsic structure of networks (\eg, depthwise convolution)
and the runtime implementation of the code framework.
The speedup of concurrent running with different networks on ImageNet
is shown in Figure~\ref{latency_comp}.

In Table~\ref{runtime}, we also show the average cosine similarity of 
outputs of different small networks.
There is no obvious relationship
between the similarity of outputs and the gain for different networks.
For example, (ResNeXt-19, $S=4$) has similarity $0.89$ and gain $1.55\%$ in accuracy, while
(EfficientNet-B0, $S=4$) has similarity $0.79$ and gain $0.65\%$ in accuracy.
(EfficientNet-B3, $S=4$) with similarity $0.85$ also has a larger gain $1.48\%$
than (EfficientNet-B0, $S=4$).
Lower similarity scores do not always mean more gains in accuracy.

\begin{figure}[tbp]
	\centering
  	\subfloat[Top-1 Accuracy\label{fig:sub1}]{%
	\includegraphics[width=0.48\linewidth]{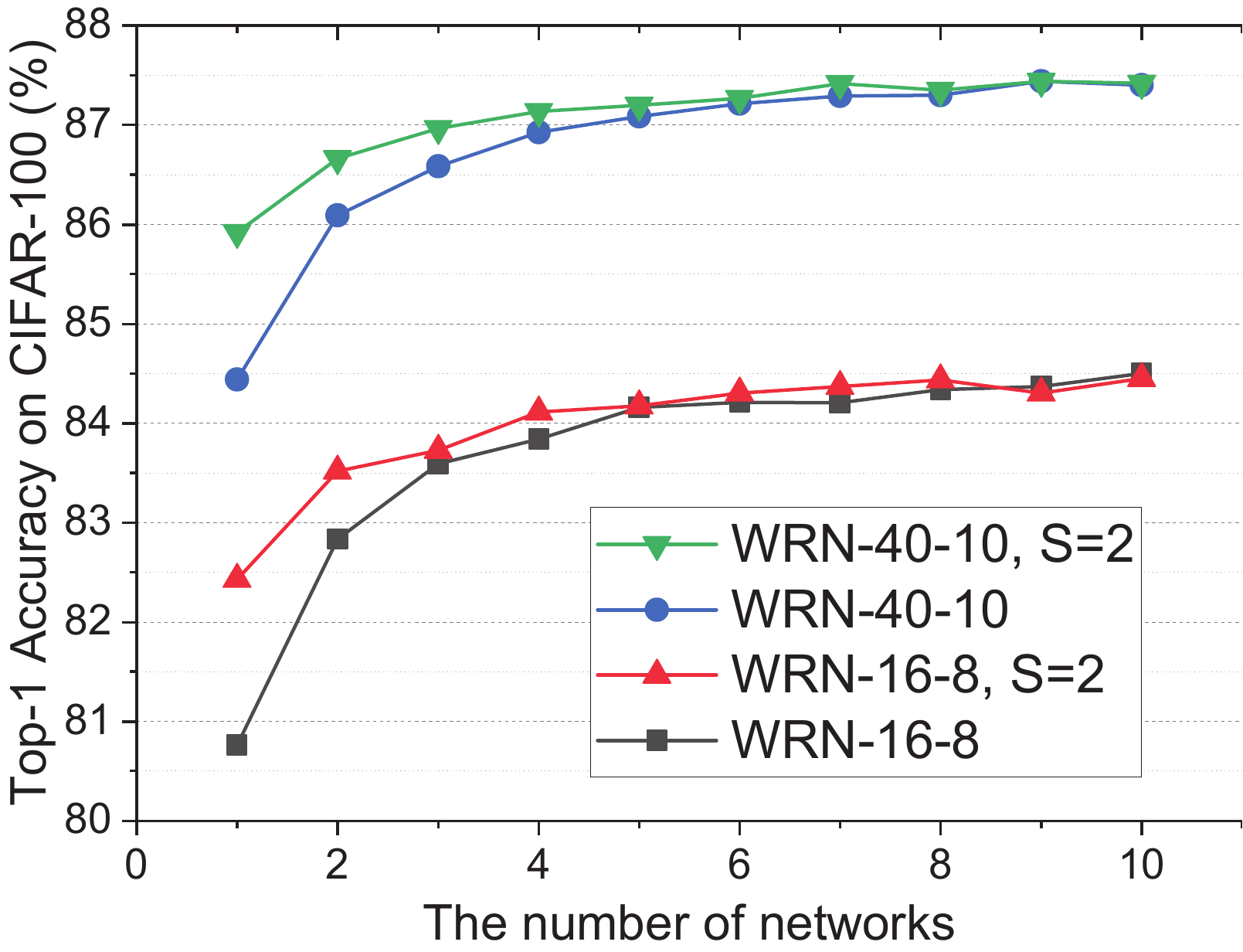}}
	\subfloat[Top-5 Accuracy\label{fig:sub2}]{%
	\includegraphics[width=0.48\linewidth]{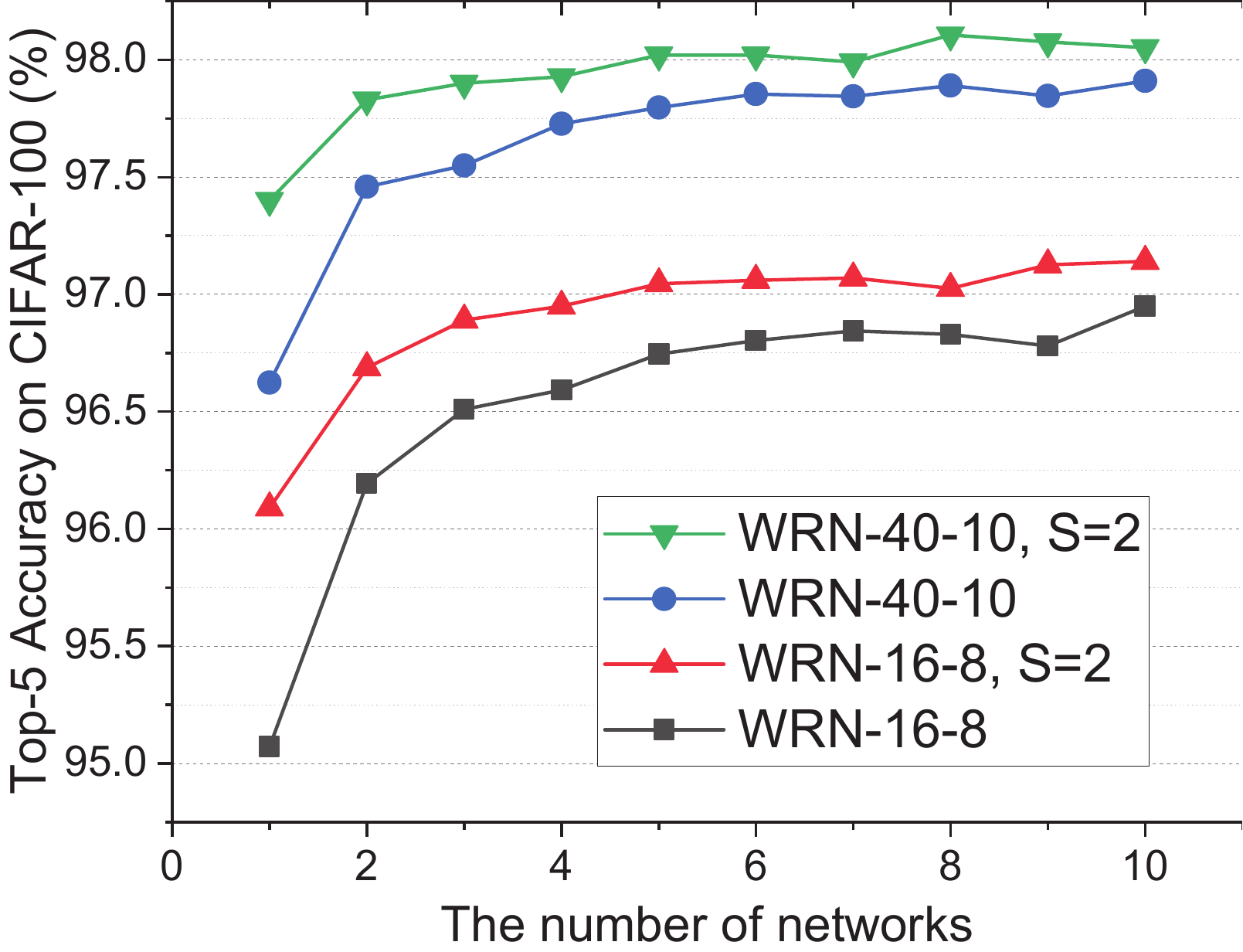}}	
	
	\caption{Ensemble of ensemble models.
		\{WRN, $S=2$\} is treated as a single model.
	The left is top-1 accuracy, the right is top-5 accuracy.}
	\label{fig_enen}
\end{figure}

\paragraph{Ensemble of ensemble models}
The ensemble of divided networks can also achieve better
accuracy-efficiency trade-offs (better performance with roughly the same parameters)
than the original single model.
The testing results are shown in Figure~\ref{fig_enen}.
For Top-1 accuracy, the ensemble of ensemble models obtains
better performance than single models when the number of
networks is relatively small. Then the former also reaches
the saturation point (plateau) faster than the latter.
As for top-5 accuracy, the ensemble of ensemble models
always achieves higher accuracy than single models.
This shows the robustness of the ensemble of ensemble models.
Like other dimensions of model scaling, \ie, width, depth, and resolution,
increasing the number of networks will also saturate.
Otherwise, we will get a perfect model by increasing the number of
networks; it is not practical.

\section{Dividing for object detection}
We discuss the dividing and co-training strategies for
image classification models in the above context.
From the experiments on image classification, we can see that dividing and ensemble 
contributes most to help the system achieve better performance
with similar computation costs.
In this section, we will explore how dividing and ensemble
work on another fundamental computer vision task --- object detection.

\begin{figure}[t]
	\centering
	\includegraphics[width=\linewidth]{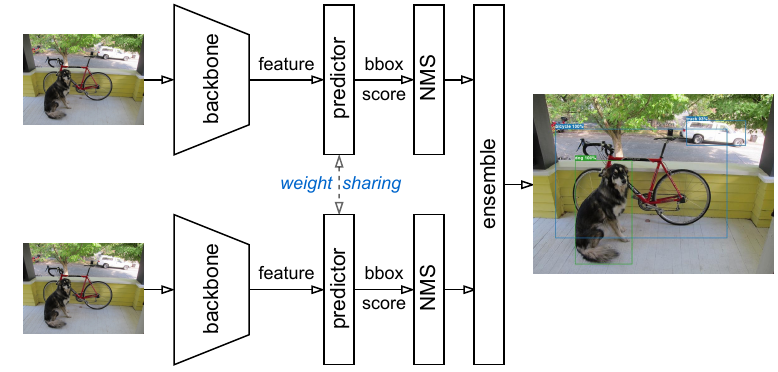}
	\caption{The overall architecture of the detection system with dividing.
		The two different backbones are obtained from dividing a large network.
	}
	\label{detection}
\end{figure}

The overall architecture of our detection system is shown in Figure~\ref{detection}.
Without loss of generality, we choose SSD~\cite{2016_ssd} as the object detector and
replace its original backbone (feature extractor) ---
VGG16~\cite{2014_VGGNet} with models in Table.~\ref{tab_imagenet_results}.
We then divide the backbone into $S$ small networks, typically, $S=2$.
For simplicity and compatibility with the bounding box regression process of SSD,
the detection predictors --- extra feature layers and classifiers in SSD
share the same weights.
After Non-Maximum Suppression (NMS),
we obtain the predicted bounding boxes and their associated confidence scores.
Furthermore, these predictions from different small detection models are composed
to get the final predicted bounding boxes and corresponding confidence scores.
To fully utilize the information provided by different detection models,
the ensemble method we adopt here is weighted boxes fusion (WBF)~\cite{2021_wbf}.
Unlike NMS-based methods~\cite{GarciaH20,2017_softnms}
which will simply abandon part of the predictions,
WBF utilizes all predicted boxes of a certain object from different models
to construct a more accurate average box based on the confidence scores of boxes.
Such dividing and ensemble methods can be easily extended to other object detectors
like YOLO~\cite{2016_YOLO}.

\begin{table*}[tbp]
	\centering
	\small
	\caption{Results of SSD300 on COCO \texttt{val2017} with dividing and ensemble.
	Average Precision (AP) and Average Recall (AR) are reported.}
	\label{tab:coco}
	\resizebox{0.86\linewidth}{!}{		
		\begin{tabular}{l|c|c|ccc|ccc|ccc|ccc}
		\toprule
		\rowcolor{Gray}
		~
		& ~
		& ~
		& \multicolumn{3}{c|}{\scriptsize{AP, IoU:}}
		& \multicolumn{3}{c|}{\scriptsize{AP, Area:}}
		& \multicolumn{3}{c|}{\scriptsize{AR, \#Dets:}}
		& \multicolumn{3}{c}{\scriptsize{AR, Area:}}  \\
		
		\rowcolor{Gray}
		\multirow{-2}{*}{Backbone}
		& \multirow{-2}{*}{MParams}
		& \multirow{-2}{*}{GFLOPs}
		& 0.5:0.95 & 0.5 & 0.75 
		& S & M & L 
		& 1 & 10 & 100 
		& S & M & L		\\
		\midrule

		ResNet-50~\cite{2020_deeplearningexamples}
		& 23.0 & 42.3
		& 25.0 & 42.3 & 25.7
		& 7.6  & 26.9 & 39.9
		& 23.7 & 34.2 & 35.8
		& 11.8 & 39.4 & 54.8		\\
		\midrule

		WRN-50-2
		& 39.3 & 48.1
		& \textbf{30.3} & \textbf{49.7} & \textbf{31.7}
		& \textbf{10.9} & \textbf{34.1} & 45.5
		& 27.0 & 39.5 & 41.2
		& 16.2 & 46.9 & 58.9		\\

		WRN-50-2, $S=2$
		& 31.7 & 45.3
		& 29.9 & 49.3 & 31.2
		& 10.5 & 33.3 & \textbf{45.7}
		& 27.0 & \textbf{39.9} & \textbf{42.1}
		& \textbf{17.1} & \textbf{47.6} & \textbf{60.9}		\\
		\midrule

		WRN-50-3
		& 64.1 & 86.8
		& 30.7 & 51.2 & 32.0
		& 11.1 & 34.8 & 45.9
		& 27.2 & 39.6 & 41.4
		& 16.5 & 46.5 & 59.9		\\

		WRN-50-3, $S=2$
		& 64.3 & 96.3
		& \textbf{31.6} & \textbf{51.5} & \textbf{33.0}
		& \textbf{11.9} & \textbf{35.6} & \textbf{47.3}
		& \textbf{28.0} & \textbf{41.2} & \textbf{43.3}
		& \textbf{18.5} & \textbf{49.2} & \textbf{61.5}		\\
		\midrule	

		ResNeXt-101,~64$\times$4d
		& 68.9 & 90.1
		& 32.6 & 53.0 & 34.1
		& 11.9 & 36.6 & 49.6
		& 28.2 & 41.0 & 42.8
		& 17.8 & 48.4 & 62.4		\\
		
		ResNeXt-101,~64$\times$4d,~$S=2$
		& 69.8 & 100.5
		& \textbf{34.1} & \textbf{54.7} & \textbf{35.8}
		& \textbf{13.5} & \textbf{38.5} & \textbf{52.2}
		& \textbf{29.3} & \textbf{43.0} & \textbf{45.2}
		& \textbf{20.4} & \textbf{51.0} & \textbf{65.5}		\\
		\midrule

		\end{tabular}}

\end{table*}

Experiments on object detection are done with SSD300 on MS COCO~\cite{2014_COCO} dataset.
The input image is resized to 300$\times$300.
Following training schedule of the open-source implementation
of NVIDIA~\cite{2020_deeplearningexamples},
we train SSD300 with different backbones on COCO \texttt{train2017} set
for 65 epochs.
The COCO \texttt{train2017} set contains 118K RGB images.
The initial learning rate is $0.0026$ and is decayed by 10 at epoch 43 and 54.
The optimization algorithm is SGD with a momentum of 0.9.
The value of weight decay is \textit{5e-4}.
After training, the model is validated on COCO \texttt{val2017} set,
which contains 5K images.
The backbones are all pre-trained on ImageNet as shown in 
Table.~\ref{tab_imagenet_results}.
No test-time augmentations are used.

The results of SSD300 with dividing and ensemble are shown in Table.~\ref{tab:coco}.
The dividing strategy also works for SSD on the object detection task.
When the backbone is large, dividing and ensemble can bring significant improvements of
AP and AR. Especially for the recall of objects with large areas,
dividing and ensemble achieve more than +3\% AR for (ResNeXt-101,~64$\times$4d,~$S=2$)
compared to its counterpart.
Due to multiple runs of predictors as Figure~\ref{detection} shows,
the FLOPs of the whole system will generally increase after dividing.
The numbers of parameters are roughly similar.

\section{Conclusion and future work}
\noindent
In this paper,
we discuss the accuracy-efficiency trade-offs
of increasing the number of networks and demonstrate
it is better than purely increasing the depth or width of networks.
Along this process, a simple yet effective method ---
dividing and co-training is proposed to enhance the performance
of a single large model.
This work potentially introduces some interesting topics in neural network
engineering, \eg,
designing a flexible framework for asynchronous training of multiple models,
more complex deep ensemble and co-training methods,
multiple models with different modalities, introducing the idea to NAS
(given a specific constrain of FLOPs,
how can we design one or several models to get the best performance).

\section*{Acknowledgments}
Thanks Boxi Wu, Yuejin Li, and all reviewers
for their technical support and helpful discussions.

{\appendix[Details about dividing a large network]
$S$ is the number of small networks after dividing.
\paragraph{ResNet}
For CIFAR-10 and CIFAR-100,
the numbers of input channels of the three blocks are:
\begin{align*}
\text{original}: \quad &[16, 32, 64], \\
S=2: \quad &[12, 24, 48], \\
S=4: \quad &[8, 16, 32].	
\end{align*}

\paragraph{SE-ResNet}
The reduction ratio in the SE module keeps unchanged.
Other settings are the same as ResNet.

\paragraph{EfficientNet}
The numbers of output channels of the first conv. layer and
blocks in EfficientNet baseline are:
\begin{align*}
\text{original}: \quad &[32, 16, 24, 40, 80, 112, 192, 320, 1280], \\
S=2: \quad &[24, 12, 16, 24, 56, 80, 136, 224, 920], \\
S=4: \quad &[16, 12, 16, 20, 40, 56, 96, 160, 640].	
\end{align*}

\paragraph{WRN}
Suppose the widen factor is $w$,
the new widen factor $w^\star$ after dividing is:
$
w^\star = \max( \lfloor \frac{w}{\sqrt{S}} + 0.4 \rfloor, 1.0).
$

\paragraph{ResNeXt}
Suppose original \textit{cardinality} (groups in convolution) is $d$, 
new \textit{cardinality} $d^\star$ is:
$
d^\star = \max( \lfloor \frac{d}{S}\rfloor, 1.0).
$

\paragraph{Shake-Shake}
For Shake-Shake 26 2$\times$96d, the numbers of output channels of the first convolutional 
layer and three blocks are:
\begin{align*}
\text{original}: \quad  [16, 96, 96\times 2, 96 \times 4], \\
S=2: \quad [16, 64, 64\times 2, 64 \times 4], \\
S=4: \quad [16, 48, 48\times 2, 48 \times 4].	
\end{align*}	

\paragraph{DenseNet}
Suppose the growth rate of DenseNet is $g_{\text{dense}}$,
the new growth rate after dividing is
$
g_{\text{dense}}^\star = \frac{1}{2} \times
\lfloor2 \times \frac{g_{\text{dense}}}{\sqrt{S}} \rfloor.
$

\paragraph{PyramidNet~+~ShakeDrop}
Suppose the additional rate of PyramidNet and final drop probability of ShakeDrop
is $g_{\text{pyramid}}$ and $p_\text{shake}$, respectively, we divide them as:
$
g_{\text{pyramid}}^\star = \frac{g_{\text{pyramid}}}{\sqrt{S}},
p_\text{shake}^\star = \frac{p_\text{shake}}{\sqrt{S}}.
$
To pursue better performance, we do not divide the base channel of PyramidNet
on CIFAR since it is small --- 16.

}

\bibliographystyle{IEEEtran}
\bibliography{egbib}


\begin{IEEEbiography}[{\includegraphics
		[width=1in,height=1.25in,clip,
		keepaspectratio]{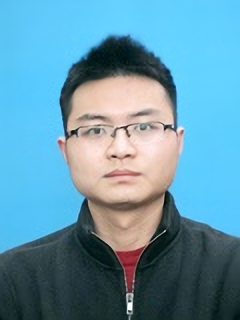}}]
	{Shuai Zhao}
received the Master degree in Computer
Science from Zhejiang University, Hangzhou, China in 2020,
and the Bachelor Degree in Automation from Huazhong University of Science \& Technology,
Wuhan, China in 2017.
His research interests include computer vision and machine learning.
\end{IEEEbiography}
\begin{IEEEbiography}[{\includegraphics
		[width=1in,height=1.25in,clip,
		keepaspectratio]{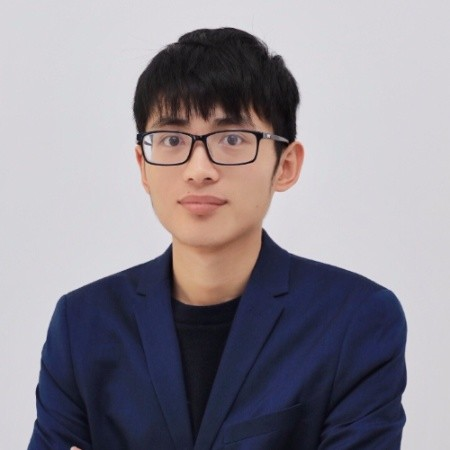}}]
	{Liguang Zhou}
received the B.Eng. degree in Electrical Engineering and Automation from China Jiliang University, Hangzhou, China,
in 2016, and he is pursuing the Ph.D. degree in Computer Information Engineering at The Chinese of Hong Kong, Shenzhen, China.

His research interests include robotic vision, scene understanding, and computational photography.
\end{IEEEbiography}
\begin{IEEEbiography}[{\includegraphics
		[width=1in,height=1.25in,clip,
		keepaspectratio]{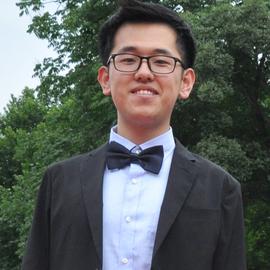}}]
	{Wenxiao Wang}
is an assistant professor, School of Software Technology at Zhejiang University, China. He received the Ph.D. degree in computer science and technology from Zhejiang University in 2022. His research interests include deep learning and computer vision.
\end{IEEEbiography}

\begin{IEEEbiography}[{\includegraphics
		[width=1in,height=1.25in,clip,
		keepaspectratio]{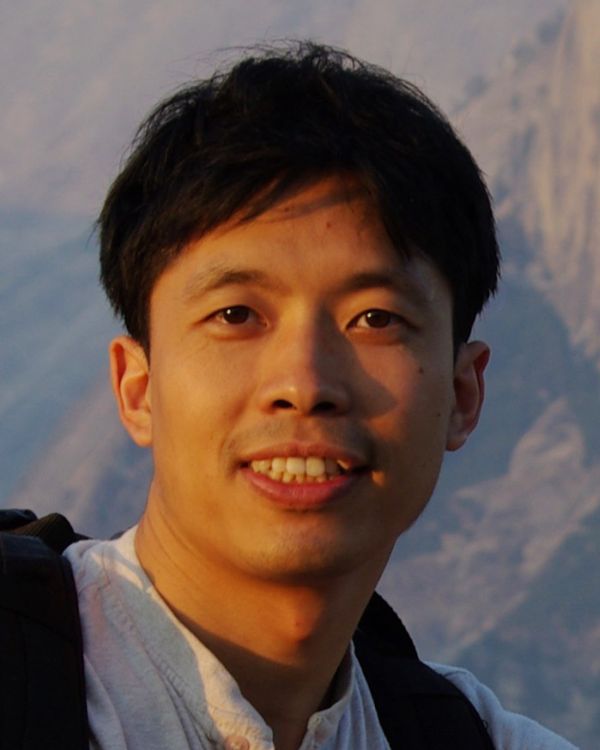}}]
	{Deng Cai}
is currently a full professor in the College
of Computer Science at Zhejiang University, China.
He received the Ph.D. degree from University of
Illinois at Urbana Champaign. His research interests
include machine learning, computer vision, data
mining and information retrieval.
\end{IEEEbiography}
\begin{IEEEbiography}[{\includegraphics
		[width=1in,height=1.25in,clip,
		keepaspectratio]{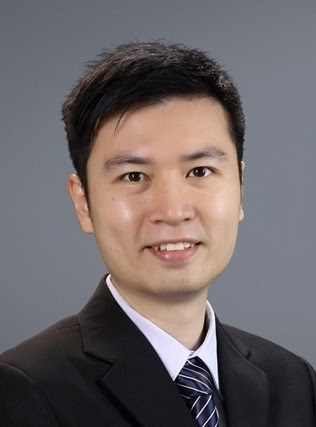}}]
	{Tin Lun LAM (Senior Member, IEEE)}
received the B.Eng. (First Class Hons.) and Ph.D. degrees in robotics and automation from the Chinese University of Hong Kong, Hong Kong, in 2006 and 2010, respectively. 
He is currently an Assistant Professor with the Chinese University of Hong Kong, Shenzhen, China, the Executive Deputy Director of the National-Local Joint Engineering Laboratory of Robotics and Intelligent Manufacturing, and the Director of Center for the Intelligent Robots, Shenzhen Institute of Artificial Intelligence and Robotics for Society. He has authored or coauthored two monographs and more than 50 research papers in top-tier international journals and conference proceedings in robotics and AI [IEEE TRANSACTIONS ON ROBOTICS, Journal of Field Robotics, IEEE/ASME TRANSACTIONS ON MECHATRONICS (T-MECH), IEEE ROBOTICS AND AUTOMATION LETTERS, IEEE International Conference on Robotics and Automation, and IEEE/RSJ International Conference on Intelligent Robots and Systems (IROS)]. He holds more than 70 patents. His research interests include multirobot systems, field robotics, and collaborative robotics. 
Dr. Lam received an IEEE/ASME T-MECH Best Paper Award in 2011 and the IEEE/RSJ IROS Best Paper Award on Robot Mechanisms and Design in 2020.
\end{IEEEbiography}
\begin{IEEEbiography}[{\includegraphics
		[width=1in,height=1.25in,clip,
		keepaspectratio]{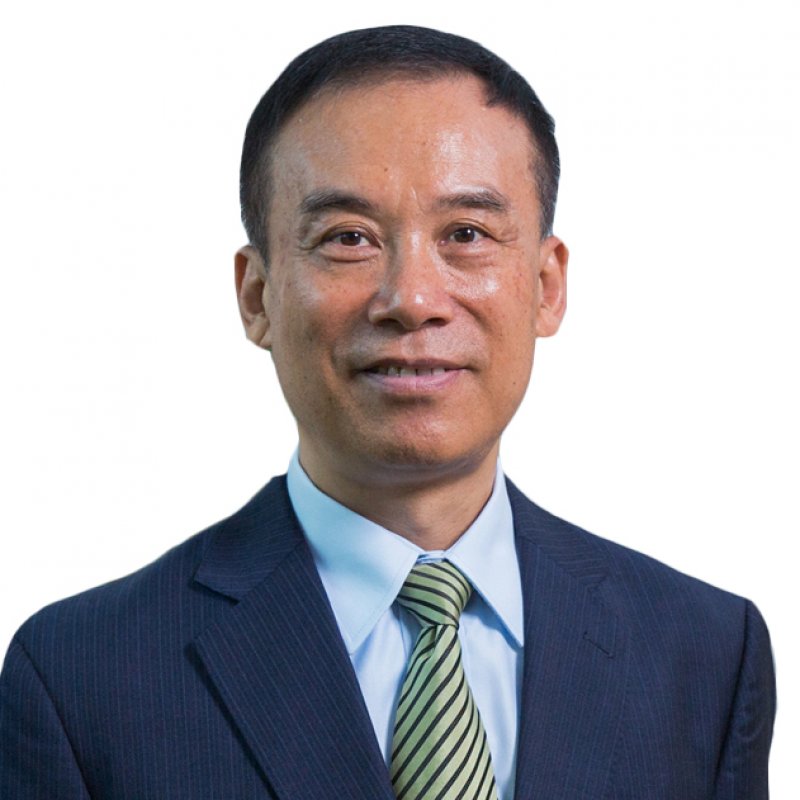}}]
	{Yangsheng Xu (Fellow, IEEE)}
received the B.S.E. and M.S.E.
degrees from Zhejiang University, Hangzhou, China,
in 1982 and 1984, respectively, and the Ph.D. degree
in 1989 from University of Pennsylvania, Philadelphia, PA, USA, all in robotics. 

He is currently President and Professor of the Chinese University of Hong Kong, Shenzhen, China. His research interests include robotics, intelligent systems, and electric vehicles.
\end{IEEEbiography}

\end{document}